\PassOptionsToPackage{table}{xcolor}
\documentclass{article}

\usepackage{amsmath}
\usepackage{amssymb}
\usepackage{booktabs}
\usepackage{array}
\usepackage{tabularx}
\usepackage{longtable}
\usepackage{graphicx}
\usepackage{float}
\usepackage{xcolor}
\usepackage{pifont}
\usepackage{enumitem}
\usepackage{comment}
\usepackage{makecell}
\usepackage[letterpaper,margin=1in]{geometry}
\usepackage[square,numbers]{natbib}
\usepackage{hyperref}
\usepackage[most]{tcolorbox}
\usepackage{marvosym}

\definecolor{metabg}{HTML}{F3F4F6}
\definecolor{metafg}{HTML}{1C2B33}

\definecolor{LinkBlue}{rgb}{0,0.08,0.45}
\hypersetup{
  pdftitle={InstructMove: A Text-Indispensable Benchmark for Instruction-Following Manipulation},
  pdfauthor={Mengao Zhao, Ziang Li, Chaodong Huang, Mengchen Ma, Haoyi Jiang, Yiwei Jin, Xinjie Wang, Yun Du, Xuewu Lin, Taojun Ding, Hongyu Xie, Jackson Jiang, Chunlei Yu, Kaihua Zhang, Lichao Huang, Liu Liu, Tianwei Lin, Zhizhong Su},
  colorlinks=true,
  linkcolor=LinkBlue,
  citecolor=LinkBlue,
  urlcolor=LinkBlue
}

\title{InstructMove: A Text-Indispensable Benchmark \\ for Instruction-Following Manipulation}

\author{
  \small
  \begin{tabular}{@{}c@{\hspace{1.25em}}c@{\hspace{1.25em}}c@{\hspace{1.25em}}c@{\hspace{1.25em}}c@{}}
    \multicolumn{5}{c}{Mengao Zhao\textsuperscript{1,*} \qquad Ziang Li\textsuperscript{1,*} \qquad Chaodong Huang\textsuperscript{1,*} \qquad Mengchen Ma\textsuperscript{3,1,*}} \\
    Haoyi Jiang\textsuperscript{4} & Yiwei Jin\textsuperscript{1} & Xinjie Wang\textsuperscript{1} & Yun Du\textsuperscript{1} & Xuewu Lin\textsuperscript{1} \\
    Taojun Ding\textsuperscript{1} & Hongyu Xie\textsuperscript{1} & Jackson Jiang\textsuperscript{2} & Chunlei Yu\textsuperscript{2} & Kaihua Zhang\textsuperscript{3} \\
    \multicolumn{5}{c}{Lichao Huang\textsuperscript{1} \qquad Liu Liu\textsuperscript{1} \qquad Tianwei Lin\textsuperscript{1} \qquad Zhizhong Su\textsuperscript{1,\Letter}} \\
    \multicolumn{5}{c}{\footnotesize \textsuperscript{1}Horizon Robotics \quad \textsuperscript{2}WuwenAI \quad \textsuperscript{3}Southeast University} \\
    \multicolumn{5}{c}{\footnotesize \textsuperscript{4}Huazhong University of Science and Technology} \\
    \multicolumn{5}{c}{\footnotesize \textsuperscript{*}Equal contribution \quad \Letter\,Correspondence: Zhizhong Su at \href{mailto:zhizhong.su@horizon.auto}{zhizhong.su@horizon.auto}}
  \end{tabular}
}
\date{}

\makeatletter
\renewcommand{\maketitle}{%
  \vspace*{-0.45in}%
  \begin{tcolorbox}[
    enhanced,
    frame hidden,
    colback=metabg,
    arc=8pt,
    left=0.5cm, right=0.5cm, top=0.35cm, bottom=0.35cm,
    grow to left by=1.5pt, grow to right by=1.5pt,
    before skip=0pt, after skip=0.3em,
    overlay={%
      \node[anchor=south east, xshift=-0.45cm, yshift=0.3cm] at (frame.south east)
        {\includegraphics[width=1.26cm]{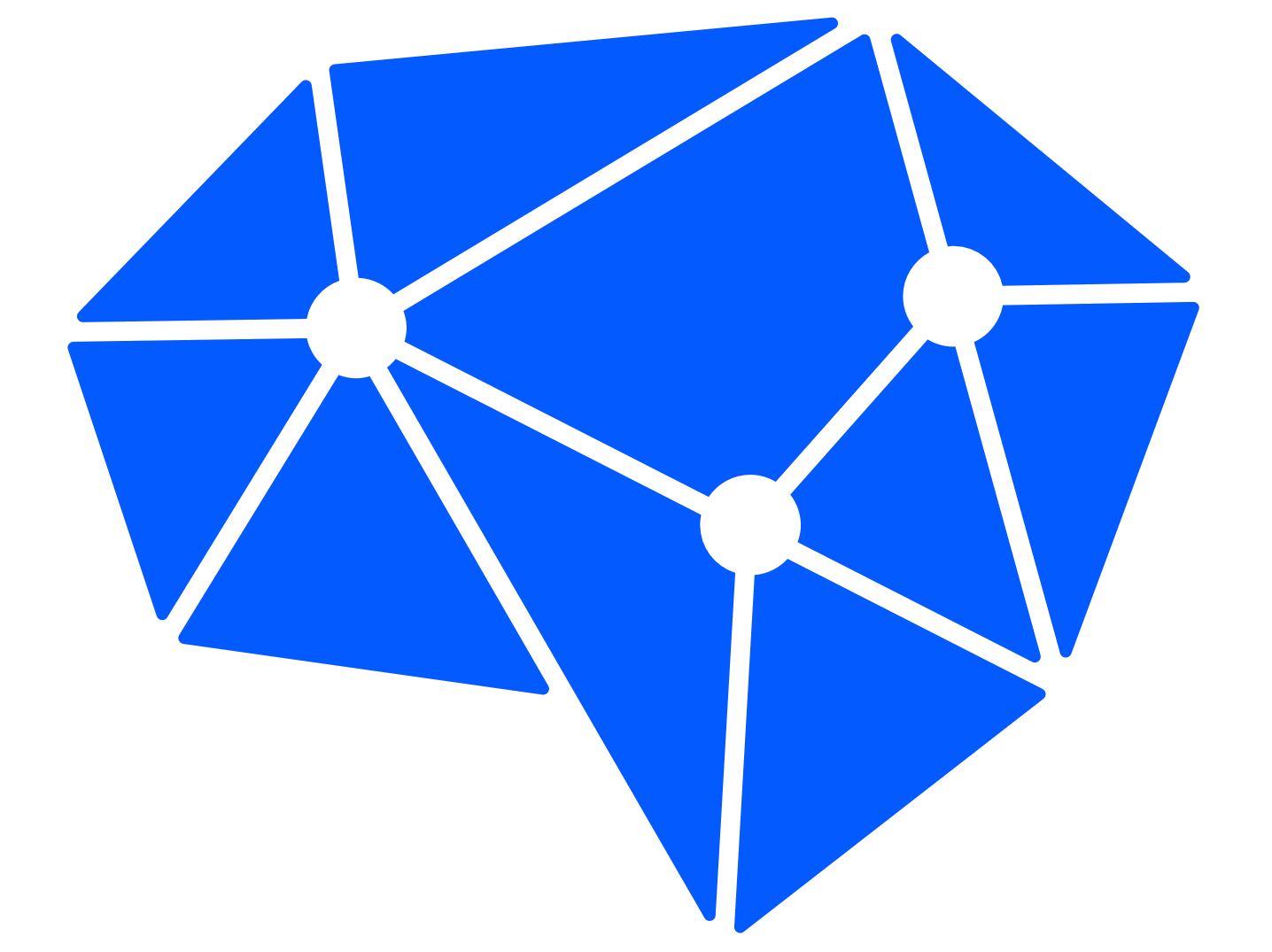}};%
    }
  ]
    \begin{center}%
      {\LARGE\bfseries \@title\par}%
      \vspace{1.3em}%
      {\@author\par}%
    \end{center}%
    \vspace{0.5em}%
    {\small\color{metafg}\noindent\textbf{Abstract.} \abstractcontent\par}%
    \vspace{0.35em}%
    {\small\noindent\textbf{Code:} \url{https://github.com/HorizonRobotics/RoboOrchardSim}\par}%
    \vspace{0.35em}%
    {\small\noindent\textbf{Keywords:} \keywordscontent\par}%
  \end{tcolorbox}%
}
\makeatother

\newcommand{\ours}{InstructMove}
\definecolor{BenchmarkGreen}{HTML}{2EB872}
\definecolor{BenchmarkRed}{HTML}{F26D78}
\definecolor{BenchmarkAmber}{HTML}{FFB347}
\newcommand{\tablemark}[1]{\raisebox{0pt}[1.45ex][0.6ex]{\makebox[1.45em][c]{\scalebox{1.22}{#1}}}}
\newcommand{\cmark}{\textcolor{BenchmarkGreen}{\ding{51}}}
\newcommand{\xmark}{\textcolor{BenchmarkRed}{\ding{55}}}
\newcommand{\pmark}{\textcolor{BenchmarkAmber}{\raisebox{0.12ex}{\scalebox{0.88}{$\blacktriangle$}}}}
\newcommand{\bcmark}{\tablemark{\cmark}}
\newcommand{\bxmark}{\tablemark{\xmark}}
\newcommand{\bpmark}{\tablemark{\pmark}}

\newcolumntype{Y}{>{\columncolor{blue!6}}c}

\begin{document}
\newcommand{\abstractcontent}{%
Vision-language-action (VLA) models have made general-purpose robot manipulation increasingly plausible by conditioning robot actions on natural-language instructions.
A key test of such generality is whether policies actually follow language instructions.
Yet many manipulation benchmarks leave this ability underdetermined: the intended object or destination is often visually salient or uniquely feasible, allowing policies to succeed without grounding the instruction.
We argue that instruction-following evaluation should be text-indispensable: multiple actions should be visually and physically plausible, while only one should be consistent with the language instruction.
We introduce \textbf{\ours}, a text-indispensable benchmark for instruction-following manipulation.
\ours{} instantiates this principle in pick-and-place scenes with semantic distractors, decomposing instruction following into category identification, attribute discrimination, spatial reasoning, and compositional pick-and-place.
\ours{} supports a train-eval protocol with \ours{} training data and held-out evaluation tasks, with additional diagnostics for language dependence.
Experiments with representative VLA policies show that \ours{} provides a controlled testbed for diagnosing visual shortcuts and that \ours{} simulation data can improve real-world instruction-following manipulation performance.
}

\newcommand{\keywordscontent}{Robot Manipulation, Instruction Following, Simulation Benchmark}

\maketitle

\begin{figure}[H]
\vspace{-0.3em}
\centering
\includegraphics[width=\linewidth]{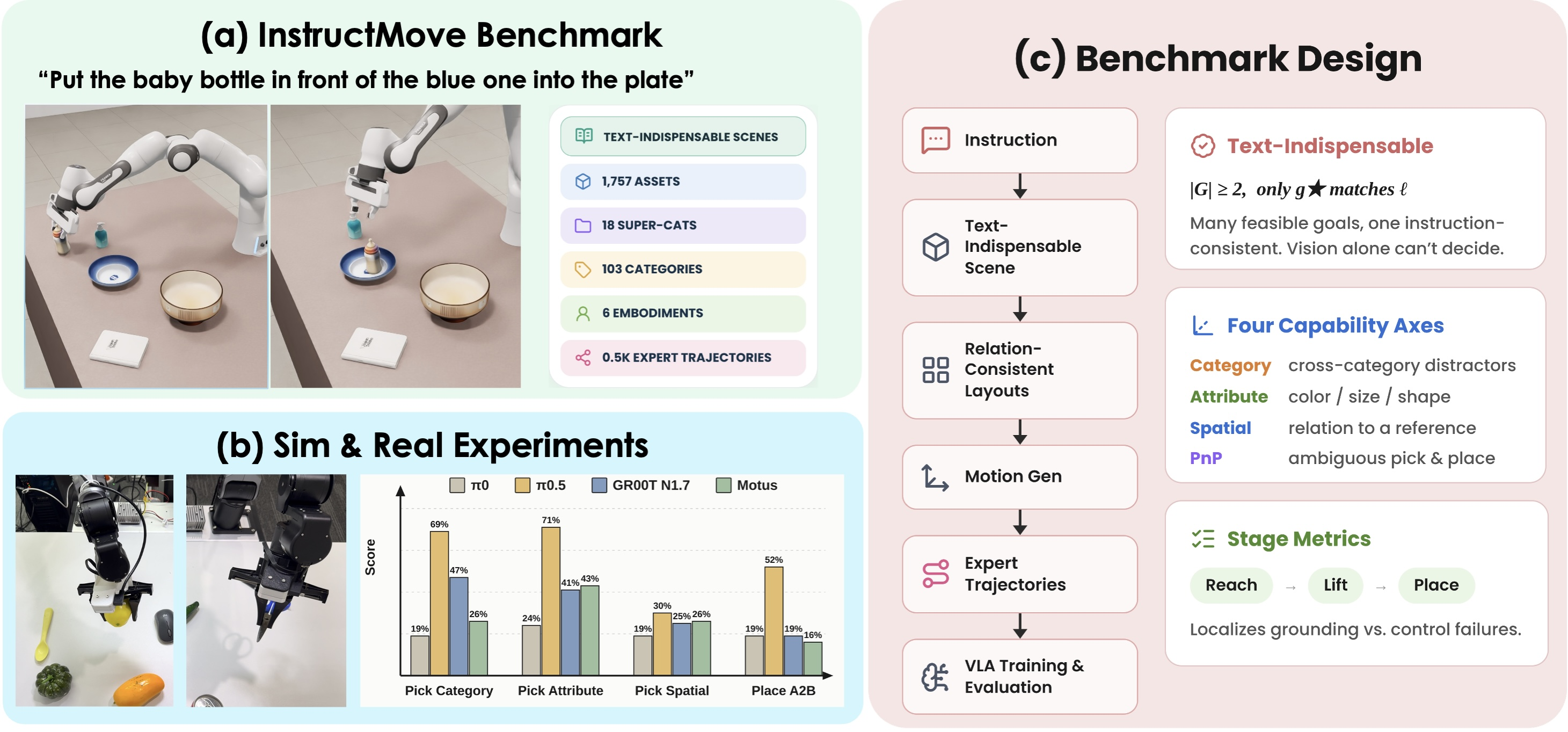}
\caption{\textbf{\ours{} is a text-indispensable benchmark for instruction-following manipulation.}
When a scene makes the target obvious, the correct pick-and-place action can be inferred from vision alone, reducing the instruction to a task identifier.
\ours{} instead introduces semantic distractors---objects and destinations that are visually plausible and physically feasible, yet are disambiguated by the instruction---making language necessary for action selection.}
\label{fig:teaser}
\end{figure}

\clearpage

\section{Introduction}

A central promise of vision-language-action (VLA) models is that natural-language instructions can determine which action a robot selects in a given scene~\citep{zitkovich2023rt2,kim2024openvla,octo2024}. For this promise to hold, different instructions should lead to different action choices within the same visual context---language must be functionally necessary, not merely descriptive. Recent benchmarks have advanced evaluation across diverse settings such as bimanual coordination, long-horizon planning, and sim-to-real transfer~\citep{liu2023libero,mees2022calvin,zhang2024vlabench,robotwin,simplerenv}. However, many evaluation scenes contain only one semantically plausible target object or placement destination. In such cases, the correct action can often be inferred from vision alone, and such visual shortcuts reduce the instruction to an auxiliary task identifier rather than a necessary grounding signal (Figure~\ref{fig:teaser}).

We argue that instruction-following evaluation should be \emph{text-indispensable}: an episode should contain multiple visually plausible and physically feasible candidate actions, while only one is uniquely consistent with the language instruction. This design makes text a necessary variable for action selection rather than optional context. It also makes visual shortcuts easier to expose: because multiple actions are physically plausible, a policy that ignores language can still accumulate non-trivial aggregate success by acting on visual saliency alone, yet stage-wise attribution keeps failures interpretable---\textit{e.g.}, reaching a distractor indicates an instruction-grounding error rather than merely a failed grasp or placement.

We introduce \ours, a text-indispensable benchmark for instruction-following manipulation. \ours{} focuses on pick-and-place because this action family provides a controlled evaluation scope while retaining the core language-grounding problem: deciding what to pick and where to place it. The benchmark makes text indispensable through semantic distractors---objects or destinations that are plausible and manipulable but inconsistent with the instruction. It decomposes instruction following into four complementary capabilities: category identification among cross-category distractors, attribute discrimination among same-category objects, spatial reasoning with respect to a reference object, and compositional pick-and-place when both the picked object and the placement destination are ambiguous.

To support controlled and reproducible evaluation, \ours{} is built on a dedicated data infrastructure comprising curated open assets with verified attribute annotations, a constraint-based layout engine that enforces relation consistency and target uniqueness, and a scalable demonstration synthesis pipeline with stage labels for reach, lift, and place events. Policies are trained on demonstrations generated by this infrastructure and evaluated on held-out tasks that test generalization across layouts, object instances, attributes, and categories. The benchmark further includes language-dependence diagnostics for measuring whether policy behavior genuinely relies on the provided instruction.

This paper makes the following contributions.

\begin{itemize}[leftmargin=*]
    \item We propose a text-indispensable benchmark for instruction-following manipulation, in which semantic distractors make language-dependent action selection necessary.
    \item We define a systematic task taxonomy covering category identification, attribute discrimination, spatial reasoning, and compositional pick-and-place.
    \item We build a scalable data infrastructure with human-curated EmbodiedGen assets~\citep{wang2025embodiedgengenerative3dworld} verified for representativeness, VLM-assisted attribute annotation, and constraint-based layout synthesis.
    \item We provide a train-eval protocol with stage-wise metrics and language-dependence diagnostics for evaluating VLA policies.
\end{itemize}

\section{Related Work}
\label{sec:related_new}

\begin{table}[t]
    \centering
    \caption{Comparison of representative manipulation and spatial vision-language benchmarks.
    \bcmark{} denotes an explicit design goal, \bpmark{} denotes partial or indirect support, and \bxmark{} denotes that the dimension is not a primary focus.
    ``ZS eval.'' denotes zero-shot policy evaluation and ``Diag.'' denotes diagnostic evaluation.
    \emph{Spatial VL} summarizes non-closed-loop spatial vision-language resources such as EmbSpatial-Bench and RoboSpatial.
    Unlike prior benchmarks, \ours{} evaluates whether language is indispensable for selecting the correct executable action among multiple visually plausible alternatives.}
    \label{tab:benchmark_comparison_new}
    \scriptsize
    \setlength{\tabcolsep}{2.15pt}
    \renewcommand{\arraystretch}{1.12}

    \resizebox{\linewidth}{!}{%
    \begin{tabular}{@{} l cccccccccc !{\color{gray!45}\vrule width 0.45pt} c @{}}
        \toprule
        \textbf{Dimension} &
        \rotatebox[origin=c]{45}{\makecell[c]{\textbf{RLBench}}} &
        \rotatebox[origin=c]{45}{\makecell[c]{\textbf{CALVIN}}} &
        \rotatebox[origin=c]{45}{\makecell[c]{\textbf{VIMA}}} &
        \rotatebox[origin=c]{45}{\makecell[c]{\textbf{LIBERO}}} &
        \rotatebox[origin=c]{45}{\makecell[c]{\textbf{VLABench}}} &
        \rotatebox[origin=c]{45}{\makecell[c]{\textbf{SimplerEnv}}} &
        \rotatebox[origin=c]{45}{\makecell[c]{\textbf{SimplerEnv-}\\\textbf{Instruct}}} &
        \rotatebox[origin=c]{45}{\makecell[c]{\textbf{RoboTwin}}} &
        \rotatebox[origin=c]{45}{\makecell[c]{\textbf{LIBERO-CF}}} &
        \rotatebox[origin=c]{45}{\makecell[c]{\textbf{Spatial}\\\textbf{VL}}} &
        \rotatebox[origin=c]{45}{\makecell[c]{\textbf{\ours{}}}} \\
        \midrule

        \multicolumn{12}{@{}l}{\cellcolor{white}\textcolor{gray!70!black}{\textbf{Visual \& Policy Setting}}} \\[1pt]
        Visual realism
            & \bxmark & \bxmark & \bxmark & \bxmark & \bpmark & \bpmark & \bpmark & \bpmark & \bxmark & \bpmark & \bpmark \\
        Language-conditioned manipulation
            & \bpmark & \bcmark & \bcmark & \bcmark & \bcmark & \bcmark & \bcmark & \bpmark & \bcmark & \bxmark & \bcmark \\
        Protocol
            & \makecell[c]{Train\\eval.}
            & \makecell[c]{Train\\eval.}
            & \makecell[c]{Train\\eval.}
            & \makecell[c]{Train\\eval.}
            & \makecell[c]{Train\\eval.}
            & \makecell[c]{ZS\\eval.}
            & \makecell[c]{ZS\\eval.}
            & \makecell[c]{Train\\eval.}
            & Diag.
            & QA
            & \textbf{\makecell[c]{Train\\eval.}} \\

        \midrule
        \multicolumn{12}{@{}l}{\cellcolor{white}\textcolor{gray!70!black}{\textbf{Grounding \& Distractor Design}}} \\[1pt]
        Semantic distractors
            & \bxmark & \bxmark & \bpmark & \bpmark & \bpmark & \bpmark & \bpmark & \bpmark & \bpmark & \bxmark & \bcmark \\
        Spatial distractors
            & \bxmark & \bxmark & \bpmark & \bpmark & \bpmark & \bxmark & \bpmark & \bpmark & \bpmark & \bcmark & \bcmark \\
        Pick-place disambiguation
            & \bpmark & \bpmark & \bpmark & \bpmark & \bcmark & \bpmark & \bpmark & \bpmark & \bpmark & \bxmark & \bcmark \\
        Feasible alternative actions
            & \bxmark & \bxmark & \bpmark & \bpmark & \bpmark & \bpmark & \bpmark & \bpmark & \bpmark & \bxmark & \bcmark \\

        \midrule
        \multicolumn{12}{@{}l}{\cellcolor{white}\textcolor{gray!70!black}{\textbf{Language-Indispensability}}} \\[1pt]
        Same-scene counterfactual instructions
            & \bxmark & \bxmark & \bxmark & \bxmark & \bxmark & \bxmark & \bxmark & \bxmark & \bcmark & \bxmark & \bcmark \\
        Text-indispensable episodes
            & \bxmark & \bxmark & \bxmark & \bxmark & \bxmark & \bxmark & \bpmark & \bxmark & \bpmark & \bxmark & \bcmark \\
        Language-dependent diagnostics
            & \bxmark & \bxmark & \bxmark & \bxmark & \bpmark & \bpmark & \bpmark & \bpmark & \bcmark & \bxmark & \bcmark \\

        \bottomrule
    \end{tabular}%
    }
\end{table}

\paragraph{Manipulation benchmarks and the missing axis.}
Language-conditioned manipulation benchmarks are now standard testbeds for robot policies. RLBench offers a large simulated suite with motion-planner demonstrations~\citep{james2019rlbench}; CALVIN emphasizes long-horizon language-conditioned control~\citep{mees2022calvin}; VIMA uses multimodal prompts and large-scale procedural tabletop tasks~\citep{jiang2023vima}; and LIBERO studies knowledge transfer and lifelong learning across language-conditioned suites~\citep{liu2023libero}. Recent benchmarks extend this scope: VLABench adds long-horizon tasks involving world knowledge, spatial relations, and vision-language-action evaluation~\citep{zhang2024vlabench}; SimplerEnv provides scalable simulated evaluation aligned with real-world policy performance~\citep{simplerenv}; SimplerEnv-Instruct targets instruction-level understanding and grounding~\citep{yang2025instructvla}; and RoboTwin scales data generation and evaluation for bimanual manipulation~\citep{chen2025robotwin}. These benchmarks support studies of task diversity, generalization, long-horizon execution, and sim-to-real transfer. However, as summarized in Table~\ref{tab:benchmark_comparison_new}, they do not primarily construct episodes where language is indispensable for choosing among multiple visually and physically plausible alternatives, even when scenes are populated with furniture and object distractors~\citep{mi2026exploratorycommunicativedeployablevisiondriven}. LIBERO-CF highlights this gap by showing that strong vision-language-action policies may follow vision-induced priors under counterfactual instructions instead of grounding the provided language~\citep{fang2026liberocf}. In contrast, \ours{} builds this axis into the benchmark: each episode contains multiple feasible candidate actions, and the instruction uniquely determines the correct object or destination.

\paragraph{Vision-Language-Action grounding and diagnostics.}
Recent VLA policies increasingly frame instruction following as grounding, not aggregate task success. Several methods preserve instruction content during action generation, add object-level supervision, or use explicit grounding intermediates such as boxes, points, masks, or spatial priors before action decoding~\citep{yang2025instructvla,physicalintelligence2025pi05,huang2025roboground,qu2025spatialvla,chen2025internvla}. Spatial reasoning resources such as EmbSpatial-Bench and RoboSpatial evaluate spatial understanding for large vision-language models, but they are not closed-loop manipulation benchmarks where spatial phrases directly determine robot behavior~\citep{du2024embspatial,song2025robospatial}. As a result, object semantics, spatial relations, and control failures are often conflated: a policy may succeed because the target is visually obvious, or fail because of grasping rather than language grounding. \ours{} turns this ambiguity into benchmark structure. It tests category grounding with cross-category distractors, attribute grounding within same-category sets, spatial grounding through relation-consistent layouts, and compositional pick-and-place through joint object and destination disambiguation. With same-scene counterfactual instructions and stage-wise metrics, \ours{} evaluates whether a policy relies on language and whether failures stem from grounding or low-level control.

\section{\ours{} Benchmark Design}
\label{sec:benchmark_new}
\ours{} turns text-indispensable instruction following into a controlled semantic-spatial pick-and-place benchmark. Each episode contains multiple feasible candidate goals, while the language instruction uniquely specifies the correct object or destination. This section formalizes the episode objective, describes the pipeline for instantiating text-indispensable scenes and demonstrations, defines the capability suites, and introduces the train-eval protocol with stage-wise metrics.

\begin{figure}[tb]
    \centering
    \includegraphics[width=\linewidth]{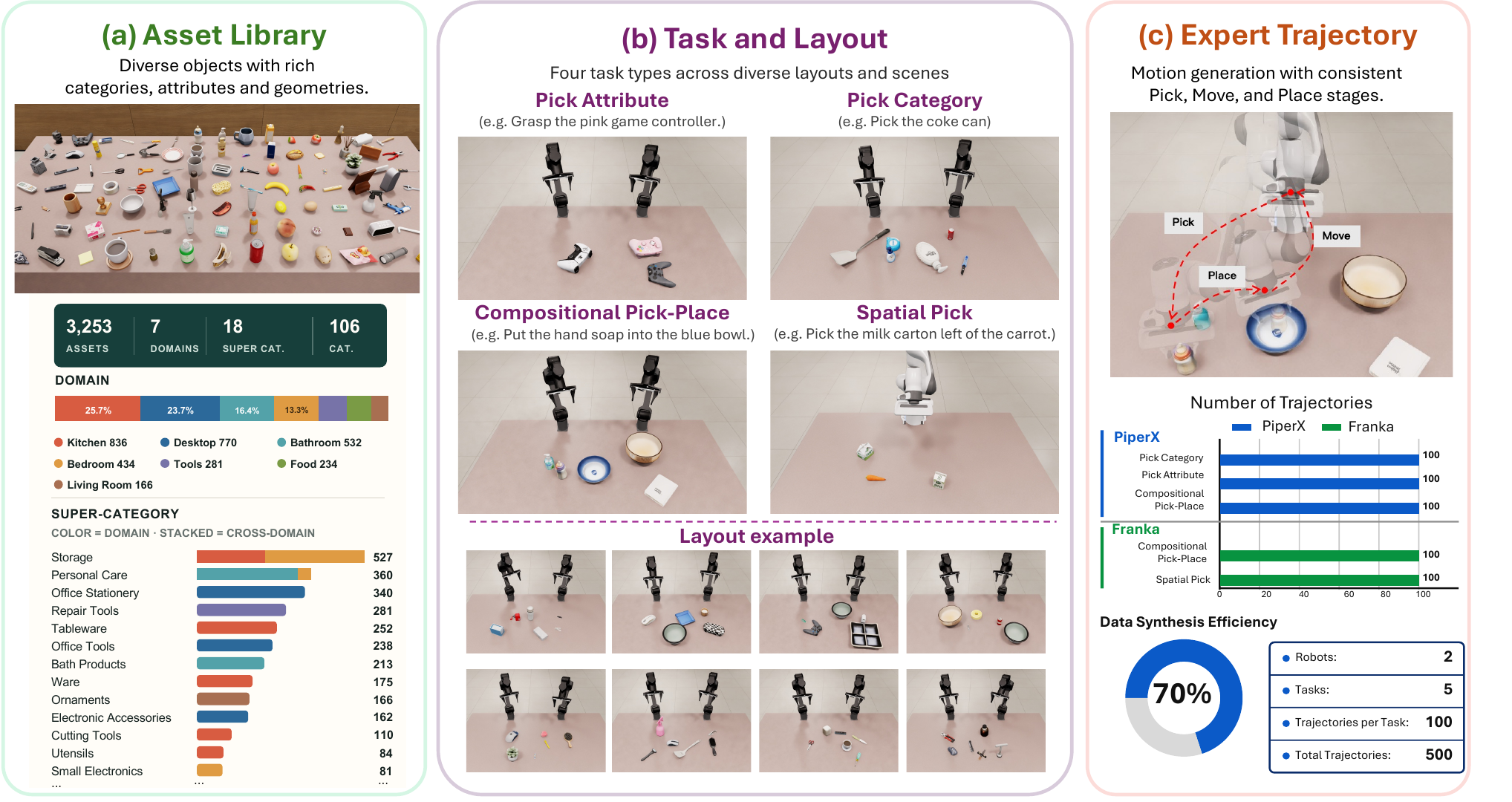}
    \caption{Overview of the InstructMove benchmark instantiation pipeline. The benchmark is built through three aligned components: a validated asset library, task templates with constraint-based scene/layout generation across four task families, and expert trajectory synthesis for generating training demonstrations.}
    \label{fig:pipeline}
\end{figure}

\subsection{Problem Formulation}

We formulate instruction-following manipulation as language-conditioned control.
At each timestep, a policy \(\pi_\theta\) receives an instruction \(\ell\) and observations \(o_t\), and produces actions
\begin{equation}
    a_t \sim \pi_\theta(a_t \mid o_{\leq t},\, \ell)
    \label{eq:policy}
\end{equation}
over a finite horizon \(T\). The episode succeeds only if the resulting trajectory completes the goal specified by \(\ell\).
Each episode is defined by a tuple
\begin{equation}
    e = (s_0,\; \mathcal{O},\; \mathcal{D},\; \ell,\; g^\star,\; \mathcal{G}),
    \label{eq:episode}
\end{equation}
where \(s_0\) denotes the initial state, \(\mathcal{O}\) the manipulable objects, \(\mathcal{D}\) the placement destinations when applicable, \(\ell\) the instruction, \(g^\star\) the instruction-consistent goal, and \(\mathcal{G}\) the set of plausible candidate goals. 
For pick-only tasks, goals specify the lifted object; for pick-and-place, they also specify the destination. Formally, text-indispensability requires
\begin{equation}
    |\mathcal{G}| \geq 2, \quad
    g^\star \in \mathcal{G}, \quad
    \text{only } g^\star \text{ matches } \ell .
    \label{eq:text_indisp}
\end{equation}
Thus, each scene contains multiple physically actionable candidate goals, but only one is instruction-consistent, so visual feasibility alone cannot identify the target. Semantic or spatial distractors expose grounding failures when the policy reaches an instruction-inconsistent candidate, while failed grasping, lifting, or placement after reaching the correct target indicates control failure. Accordingly, \ours{} reports both final success and instruction-conditioned stage-wise metrics.

\subsection{Benchmark Instantiation Pipeline}
\label{subsec:benchmark_instantiation}
All simulated benchmark environments are developed using NVIDIA Isaac Lab~\citep{mittal2025isaaclab} and NVIDIA Isaac Sim~\citep{nvidia2026isaacsim}.
As shown in Figure~\ref{fig:pipeline}, \ours{} instantiates text-indispensable instruction-following manipulation data through three aligned components: an asset library, a task-and-layout library, and an expert trajectory dataset. The asset library provides the semantic and physical basis for task construction, built from a curated collection of 3,253 objects from which \ours{} selects 1,757 for benchmark instantiation; these objects span diverse household categories, attributes, and geometric variation, enabling instructions to refer to targets through category-, attribute-, and relation-relevant cues, and the library remains extensible for constructing new benchmark instances or expanding the object pool under the same semantic-spatial pipeline. Built on top of these assets, the task-and-layout library defines the benchmark episodes across four task families---category grounding, attribute grounding, spatial grounding, and compositional pick-and-place---instantiating scenes with diverse object layouts and environment variations such that multiple candidate objects or goals remain physically plausible, while only one is consistent with the instruction.

From these instantiated scenes, \ours{} further constructs an expert trajectory dataset with consistent manipulation stages, connecting language instructions, scene configurations, and robot actions into executable pick-and-place trajectories that supervise policy training and provide a unified basis for evaluation. Together, the three components align benchmark construction, data generation, and downstream evaluation within a shared semantic-spatial setting.

\subsection{Capability Axes and Task Suites}
\ours{} evaluates semantic-spatial grounding under controlled ambiguity. Its four task families test category-level object selection, attribute discrimination, spatial relation understanding, and compositional pick-and-place reasoning. Each family uses a different instruction cue to refer to the target, such as category, attribute, spatial relation, or placement destination. Distractors are constructed around the same cue, making unrelated shortcuts unreliable. 

\textbf{Pick Category} evaluates whether a policy can identify the object category. Each scene contains one target object and five cross-category distractors, and the policy must reach, grasp, and lift the instructed category. Instructions follow templates such as ``grasp the \{object description\}.'' Evaluation spans three splits: \emph{seen}, in which categories and assets appear during training but test seeds are held out; \emph{unseen-category}, in which both target and distractor categories are held out; and \emph{unseen-instance}, in which categories are seen but specific object instances are held out.

\textbf{Pick Attribute} evaluates attribute grounding within an object category. Each scene contains one target and two same-category distractors that differ in attributes such as color, size, shape, or material. Instructions follow templates such as ``lift the \{object attribute\} \{object category\}.'' Because all candidates share a category, the policy must ground the attribute phrase rather than rely on object-type recognition.

\textbf{Pick Spatial} evaluates relational grounding. Each scene contains a reference object, a target object, and a same-category distractor. The target and distractor belong to the same object category, preventing category identity from revealing the target and requiring the policy to resolve the spatial relation rather than exploit an object-type shortcut. The spatial-relation vocabulary includes \emph{left of}, \emph{right of}, \emph{in front of}, \emph{behind}, \emph{near}, and \emph{far from}, all defined with respect to the robot-centric reference frame (i.e., from the robot's perspective). Instructions follow templates such as ``pick the \{object category\} \{spatial relation\} the \{reference category\}.'' Layouts are generated so that exactly one candidate satisfies the specified spatial relation.

\textbf{Compositional Pick-Place} evaluates instruction grounding across a pick-and-place sequence. Each scene contains multiple plausible pick objects and placement goals, creating ambiguity over what to pick and where to place it. Instructions follow templates such as ``pick up the \{pick object description\} and place it to \{place object description\}.'' This setting tests whether a policy can compose object selection, spatial destination grounding, and multi-stage control into a single language-conditioned manipulation behavior.

\subsection{Evaluation Protocol and Stage Metrics}
\label{sec:metrics}

\ours{} supports a train-eval protocol over the four task families. In each instantiated robot setting, policies are trained on demonstrations generated by \ours{} and evaluated on held-out tasks covered by the corresponding training mixture. This protocol measures instruction-following manipulation under controlled generalization across unseen scene configurations, object instances, attributes, and categories.
Final success can conflate language grounding with low-level control. \ours{} therefore reports stage metrics conditioned on instruction-consistent targets. For pick-only tasks, the score is
\begin{equation}
S_{\mathrm{pick}} = 0.5 \cdot \mathbb{I}[\mathrm{Reach}] + 0.5 \cdot \mathbb{I}[\mathrm{Lift}],
\end{equation}
where Reach indicates that the end-effector comes within a predefined distance of the instruction-consistent target, primarily evaluating visual grounding, while Lift indicates that the target is successfully grasped and raised, primarily evaluating manipulation control. For compositional pick-and-place, the score is
\begin{equation}
S_{\mathrm{pnp}} =
0.25 \cdot \mathbb{I}[\mathrm{ReachPick}]
+ 0.25 \cdot \mathbb{I}[\mathrm{LiftPick}]
+ 0.25 \cdot \mathbb{I}[\mathrm{ReachPlace}]
+ 0.25 \cdot \mathbb{I}[\mathrm{Place}].
\end{equation}
ReachPick and LiftPick follow the Reach and Lift definitions for the instructed object, while ReachPlace and Place measure approaching and completing the instruction-consistent placement, respectively.
These staged scores localize failures: reaching a distractor indicates an instruction-grounding failure, while reaching the correct target but failing to lift or place reveals a control failure.

\section{Experiments}
\label{sec:experiments}

\subsection{Experimental Setup}

\paragraph{Models and task coverage.}
We evaluate four representative policies, $\pi_0$~\cite{black2024pi0}, $\pi_{0.5}$~\cite{physicalintelligence2025pi05}, GR00T N1.7~\cite{bjorck2025gr00t}, and Motus~\cite{bi2025motus}, under the train-eval protocol. Training is performed in a robot-specific multi-task setting: dual-arm PiperX policies are trained on \texttt{pick\_category}, \texttt{pick\_attribute}, and \texttt{place\_a2b}, whereas single-arm Franka policies are trained on \texttt{pick\_spatial} and \texttt{place\_a2b}. Additional training recipes and failure-case analyses are provided in the supplementary material.

\paragraph{Training protocol.}
All models are fine-tuned for 5{,}000 steps on eight NVIDIA RTX 5090 GPUs. We follow the recommended settings of the corresponding official implementations, with embodiment-specific adaptations where necessary. Because the policies differ in model capacity and trainable parameter scope, we use model-specific batch sizes. Full optimization settings and batch-size configurations are provided in the supplementary material.

\paragraph{Data scale and evaluation splits.}
Each task dataset contains 100 training episodes. We intentionally use compact training sets for limited post-training, enabling adaptation to the benchmark domain while retaining the capabilities of the original pretrained model. Training episodes are balanced within each task so that object instances appear approximately uniformly. Each reported metric in Table~\ref{tab:main_results} and Table~\ref{tab:pick_category_generalization} is evaluated over 100 held-out episodes, with evaluation objects balanced in the same manner. For \texttt{pick\_category}, we report three splits: \emph{seen}, which holds out layouts and distractors while reusing training categories; \emph{unseen-instance}, which evaluates novel object instances from categories observed during training; and \emph{unseen-category}, which evaluates target categories excluded from training.

\subsection{Train-Eval Main Evaluation}

\begin{figure}[t]
    \centering
    \includegraphics[width=\linewidth]{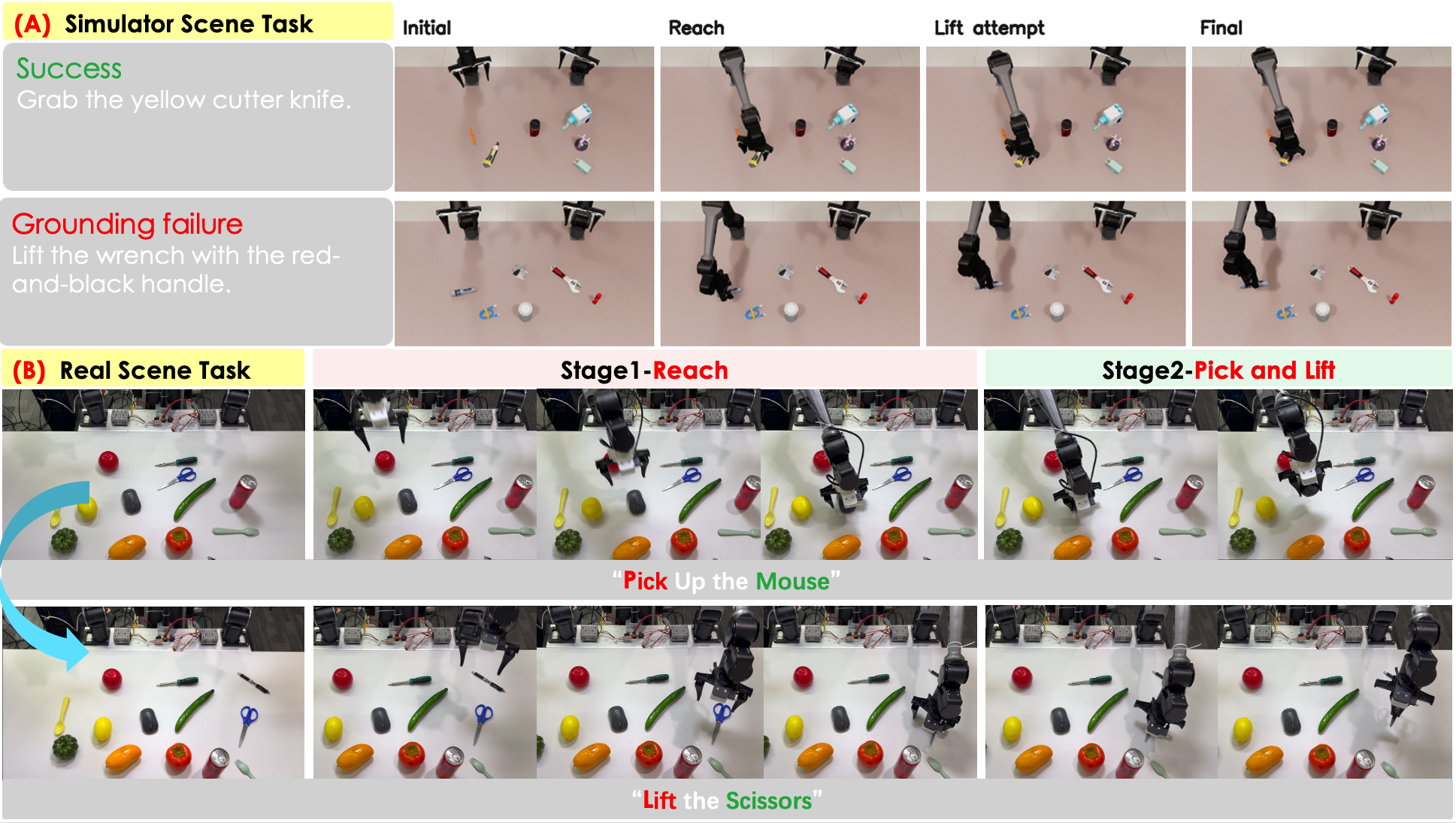}
    \caption{Qualitative examples in simulated and real evaluation cases.}
    \label{fig:fig_eval}
\end{figure}

Table~\ref{tab:main_results} reports the main train-eval results for the four policies, and Table~\ref{tab:pick_category_generalization} separately reports the held-out instance and held-out category splits for pick-category generalization. The qualitative examples in Figure~\ref{fig:fig_eval}(A) illustrate the corresponding outcomes. Benchmark stability tests show that repeated evaluations vary by no more than five percentage points, supporting the reliability of the model comparisons and performance trends reported below.

\begin{table}[h]
    \centering
    \small
    \caption{Main train-eval results after robot-specific multi-task training. The \texttt{pick\_category} and \texttt{pick\_attribute} tasks are evaluated on PiperX, and \texttt{pick\_spatial} and \texttt{place\_a2b} are evaluated on Franka. Pick-only entries are reported as Reach/Lift (\textcolor{red}{$S_{\mathrm{pick}}$}), and pick-and-place entries as ReachPick/LiftPick/ReachPlace/Place (\textcolor{red}{$S_{\mathrm{pnp}}$}); red values in parentheses denote the stage score.}
    \label{tab:main_results}
    \resizebox{\linewidth}{!}{
    \begin{tabular}{l | c c | c c}
    \toprule
    \textbf{Model}
    & \makecell[c]{\textbf{pick\_category}\\\textbf{Seen}}
    & \makecell[c]{\textbf{pick\_attribute}}
    & \makecell[c]{\textbf{pick\_spatial}}
    & \makecell[c]{\textbf{place\_a2b}} \\
    & \multicolumn{2}{>{\columncolor{gray!8}}c|}{\textit{\footnotesize PiperX}}
    & \multicolumn{2}{>{\columncolor{gray!8}}c}{\textit{\footnotesize Franka}} \\
    \midrule
    $\pi_0$
    & 0.34/0.04(\textcolor{red}{0.19})
    & 0.44/0.05(\textcolor{red}{0.24})
    & 0.36/0.02(\textcolor{red}{0.19})
    & 0.64/0.07/0.05/0.03(\textcolor{red}{0.19}) \\
    $\pi_{0.5}$
    & \textbf{0.82/0.57(\textcolor{red}{0.69})}
    & \textbf{0.81/0.62(\textcolor{red}{0.71})}
    & \textbf{0.46/0.15(\textcolor{red}{0.30})}
    & \textbf{0.88/0.48/0.37/0.36(\textcolor{red}{0.52})} \\
    GR00T N1.7
    & 0.76/0.17(\textcolor{red}{0.47})
    & 0.76/0.07(\textcolor{red}{0.41})
    & 0.50/0.01(\textcolor{red}{0.25})
    & 0.60/0.06/0.05/0.05(\textcolor{red}{0.19}) \\
    Motus
    & 0.41/0.10(\textcolor{red}{0.26})
    & 0.69/0.17(\textcolor{red}{0.43})
    & 0.51/0.02(\textcolor{red}{0.26})
    & 0.57/0.02/0.02/0.01(\textcolor{red}{0.16}) \\
    \bottomrule
    \end{tabular}
    }
\end{table}

\begin{table}[h]
    \centering
    \small
    \caption{Pick-category generalization results on PiperX for held-out instances and categories. Entries are reported as Reach/Lift (\textcolor{red}{$S_{\mathrm{pick}}$}); red values in parentheses denote the stage score.}
    \label{tab:pick_category_generalization}
    \begin{tabular}{l c c}
    \toprule
    \textbf{Model}
    & \textbf{Unseen Inst.}
    & \textbf{Unseen Cat.} \\
    \midrule
    $\pi_0$
    & 0.32/0.05(\textcolor{red}{0.18})
    & 0.34/0.05(\textcolor{red}{0.19}) \\
    $\pi_{0.5}$
    & \textbf{0.74/0.48(\textcolor{red}{0.61})}
    & \textbf{0.81/0.54(\textcolor{red}{0.67})} \\
    GR00T N1.7
    & 0.70/0.09(\textcolor{red}{0.40})
    & 0.67/0.18(\textcolor{red}{0.43}) \\
    Motus
    & 0.45/0.06(\textcolor{red}{0.26})
    & 0.63/0.09(\textcolor{red}{0.36}) \\
    \bottomrule
    \end{tabular}
\end{table}

Table~\ref{tab:main_results} shows that $\pi_{0.5}$ has the strongest target-localization ability in the semantic pick tasks. It achieves the highest Reach on both \texttt{pick\_category} and \texttt{pick\_attribute}, indicating better grounding of category and attribute instructions. GR00T N1.7 follows in Reach performance on these tasks, while $\pi_{0.5}$ also performs most robustly on compositional \texttt{place\_a2b}. $\pi_{0}$ and Motus show moderate Reach performance.

The gap between Reach and Lift further reveals differences in converting correct target localization into successful manipulation. $\pi_{0.5}$ maintains substantially higher Lift rates once the target is reached, while GR00T N1.7 reaches the instruction-consistent target relatively often but rarely completes the lift. In the Franka setting, $\pi_{0.5}$, GR00T N1.7, and Motus obtain comparable Reach rates on \texttt{pick\_spatial}, highlighting spatial-relation grounding as a central bottleneck.

Table~\ref{tab:pick_category_generalization} shows that $\pi_{0.5}$ remains close to its seen-split performance on both the unseen-instance and unseen-category splits, indicating strong generalization in category grounding. GR00T N1.7 also maintains relatively high Reach on both unseen splits, but its substantially lower Lift suggests weaker execution after correct target localization. These results suggest that policies can generalize beyond memorized object instances and categories, although reliably converting target localization into successful manipulation remains challenging.

\subsection{Counterfactual Language-Dependence Diagnostic}
\label{sec:language_ablation}

We conduct a counterfactual analysis by keeping the scene fixed and changing only the instruction. We use the $\pi_{0.5}$ results on the \texttt{pick\_category} unseen-instance split as the baseline to examine how instruction changes affect target selection and grasp initiation. We consider five conditions. \emph{Normal} uses the original instruction, while \emph{Alternative target} replaces the target description with another visible object and treats that object as the instruction-consistent target. The remaining conditions do not provide a unique valid target: \emph{Generic object} uses the underspecified description ``object,'' \emph{Empty} removes the instruction, and \emph{Absent object} names an object not present in the scene. For the first two conditions, we report target-conditioned Reach and Lift. For the other conditions, we report whether the policy reaches or lifts any object.

\begin{table}[H]
    \centering
    \small
    \caption{Counterfactual instruction-ablation results for $\pi_{0.5}$ over 100 fixed \texttt{pick\_category} unseen-instance scenes, with each scene evaluated under every instruction condition. For target-specific instructions, entries are reported as Reach/Lift (\textcolor{red}{$S_{\mathrm{pick}}$}); red values in parentheses denote the stage score. For instructions without a unique valid target, we report whether the policy reaches or lifts any object.}
    \label{tab:language_ablation}
    \resizebox{\linewidth}{!}{
    \begin{tabular}{l c c c}
    \toprule
    \textbf{Instruction condition}
    & \textbf{Target Reach/Lift} \textcolor{red}{\textbf{($S_{\mathrm{pick}}$)}}
    & \textbf{Any-object Reach}
    & \textbf{Any-object Lift} \\
    \midrule
    Normal             & 0.74/0.48(\textcolor{red}{0.61}) & --   & --   \\
    Alternative target & 0.72/0.53(\textcolor{red}{0.62}) & --   & --   \\
    \midrule
    Generic object     & --                               & 1.00 & 0.66 \\
    Empty              & --                               & 1.00 & 0.64 \\
    Absent object      & --                               & 0.99 & 0.66 \\
    \bottomrule
    \end{tabular}
    }
\end{table}

As shown in Table~\ref{tab:language_ablation}, replacing the original target with another visible object yields 0.72 Target Reach and 0.53 Target Lift, compared with 0.74 and 0.48 under the normal instruction. This result shows that the policy can use a valid language description to locate a new instruction-consistent target. However, under the generic, empty, and absent-target conditions, the policy obtains Any-object Reach rates of 0.99--1.00 and Any-object Lift rates of 0.64--0.66. These conditions run for the full evaluation horizon, whereas target-specific episodes may terminate after a successful target lift, so the any-object and target-conditioned lift rates are not directly comparable. Nevertheless, the policy frequently executes a generic grasp without a valid target, indicating that language can redirect target selection but does not fully gate whether a grasp is initiated and revealing a strong grasping prior consistent with a visual shortcut.

These results also demonstrate why \ours{} requires text-indispensable scenes and instruction-conditioned metrics. When the intended object or destination is visually salient or uniquely feasible, the policy may select the same action regardless of whether it receives a target-specific, generic, empty, or absent-target instruction. Consequently, physically successful manipulation in such scenes does not reveal whether the policy actually responds to the instruction, and actions under instructions without a valid target should not count as success.

\subsection{Real-World Transfer with Simulation Data}

\paragraph{Setup.}
We design this experiment to assess whether \ours{} simulation data transfers effectively and improves real-world manipulation performance. To isolate the effect of training data, the simulated setup mirrors the real system in robot embodiment, camera models, and mounting configuration: a dual-arm PiperX with one overhead Intel RealSense D455 and two wrist-mounted Intel RealSense D405 cameras.

We select five real-world object instances as evaluation targets and collect 200 real-world demonstrations with these same instances through ALOHA-style homologous teleoperation~\citep{zhao2023aloha}. Leveraging our large, curated asset library and scalable data-synthesis pipeline, we synthesize demonstrations for the same five object categories using a broader range of virtual object instances, providing greater intra-category diversity. We additionally apply domain randomization over lighting, tabletop textures, and robot initial poses to mitigate the sim-to-real gap. We fine-tune the same pretrained $\pi_{0.5}$ checkpoint using an identical training recipe and evaluate the resulting policies under the same real-world protocol. For the mixed-data configuration, each training batch samples simulation and real-world demonstrations at a 9:1 ratio. Each target is evaluated in 10 trials, yielding 50 real-world episodes. Each trial includes several graspable distractors selected to be visually or semantically similar, preventing the target from being identified through visual saliency alone.

\begin{table}[t]
    \centering
    \small
    \caption{Real-world Pick Category transfer with $\pi_{0.5}$ on the dual-arm PiperX. The evaluation contains 50 trials across five target objects, with at least 5 visually or semantically similar distractors per trial.}
    \label{tab:real_pick_category}
    \begin{tabular}{l c c}
    \toprule
    \textbf{Training set} & \textbf{Reach} & \textbf{Lift} \\
    \midrule
    200 real & 19/50 & 11/50 \\
    200 sim & 26/50 & 11/50 \\
    2,000 sim & 28/50 & 15/50 \\
    2,000 sim + 200 real & 34/50 & 18/50 \\
    \bottomrule
    \end{tabular}
\end{table}

\paragraph{Results.}
As shown in Table~\ref{tab:real_pick_category}, at the same data scale, the 200-sim policy improves Reach from 19/50 to 26/50 over the 200-real policy while matching its Lift at 11/50. This result suggests that the greater intra-category instance and appearance diversity of the simulation demonstrations benefits instruction-conditioned target grounding among similar distractors. Scaling the simulation data from 200 to 2,000 demonstrations yields further gains, increasing Reach from 26/50 to 28/50 and Lift from 11/50 to 15/50.

Combining 2,000 simulation demonstrations with 200 real-world demonstrations achieves the best performance, reaching 34/50 in Reach and 18/50 in Lift. Compared with simulation-only training at the same simulation scale, the real-world demonstrations provide complementary signals for visual and physical variations not fully captured in simulation. Overall, these results show that scalable simulation data can improve real-world manipulation performance, while a small amount of real-world data further strengthens transfer and reduces reliance on large-scale real-robot data collection. The qualitative examples in Figure~\ref{fig:fig_eval}(B) illustrate the corresponding real-world execution stages.

\section{Limitations}
\label{sec:limitation}

\ours{} deliberately focuses on language-grounded pick-and-place, which enables tight control over visual ambiguity, object attributes, spatial relations, and placement distractors. This focused scope does not cover longer-horizon tasks, articulated-object interaction, or manipulation in richer, less structured environments. Results should therefore be interpreted as a diagnostic of language-grounded target selection and compositional pick-and-place, rather than as a comprehensive evaluation of general-purpose manipulation. Future versions will extend the benchmark to broader task families while preserving its core principle of controlled, text-indispensable evaluation.

The benchmark also relies on synthetic assets and simulated layouts, which can introduce a domain gap relative to real deployment. We mitigate this through asset quality control, domain randomization, and real-data mixing, and our real-world experiments show that training on \ours{} simulation data can transfer to and improve real-world manipulation performance. However, the current experiments do not establish whether model rankings or absolute evaluation scores in simulation correlate with real-world performance. Future versions will expand paired simulation and real-world evaluations across models, tasks, and embodiments, with the goal of making the simulation metrics more predictive of real-world performance. These results demonstrate the utility of simulation training data but do not yet establish simulated evaluation as a reliable proxy for real-world performance.

The current train-eval protocol fine-tunes and evaluates policies on the same task families, although the evaluation episodes use held-out layouts, object instances, and categories. While this protocol measures adaptation to the benchmark tasks, task-specific fine-tuning may encourage overfitting or degrade pretrained capabilities, making it difficult to separate general instruction-following ability from adaptation to the task templates. An alternative is to provide task-agnostic training data that exposes policies to the visual and physical characteristics of the simulator without using the benchmark task definitions. Future work will compare these protocols to determine whether domain-only adaptation can better preserve pretrained capabilities while still supporting reliable evaluation in simulation.

\section{Conclusion}
\label{sec:conclusion}

We presented \ours, a benchmark for evaluating language-grounded pick-and-place under semantic distractors. It constructs semantically dense scenes where the correct target cannot be inferred from vision alone, decomposes the problem into four complementary task families, and reports stage metrics separating semantic grounding from control execution. The benchmark helps the community diagnose visual shortcuts in VLA policies and develop models that act more reliably on language. Future versions can extend this design to longer-horizon tasks, articulated objects, and richer scenes while preserving the text-indispensability principle that behavior must be grounded in the instruction.

\clearpage
\bibliographystyle{unsrtnat}
\bibliography{example}

\clearpage
\appendix
\section{Benchmark Infrastructure and Asset Pipeline}
\label{sec:supp_benchmark_infrastructure}

\subsection{Benchmark Infrastructure Details}
\label{subsec:supp_benchmark_infrastructure}

\begin{figure}[h]
\centering
\includegraphics[width=\linewidth]{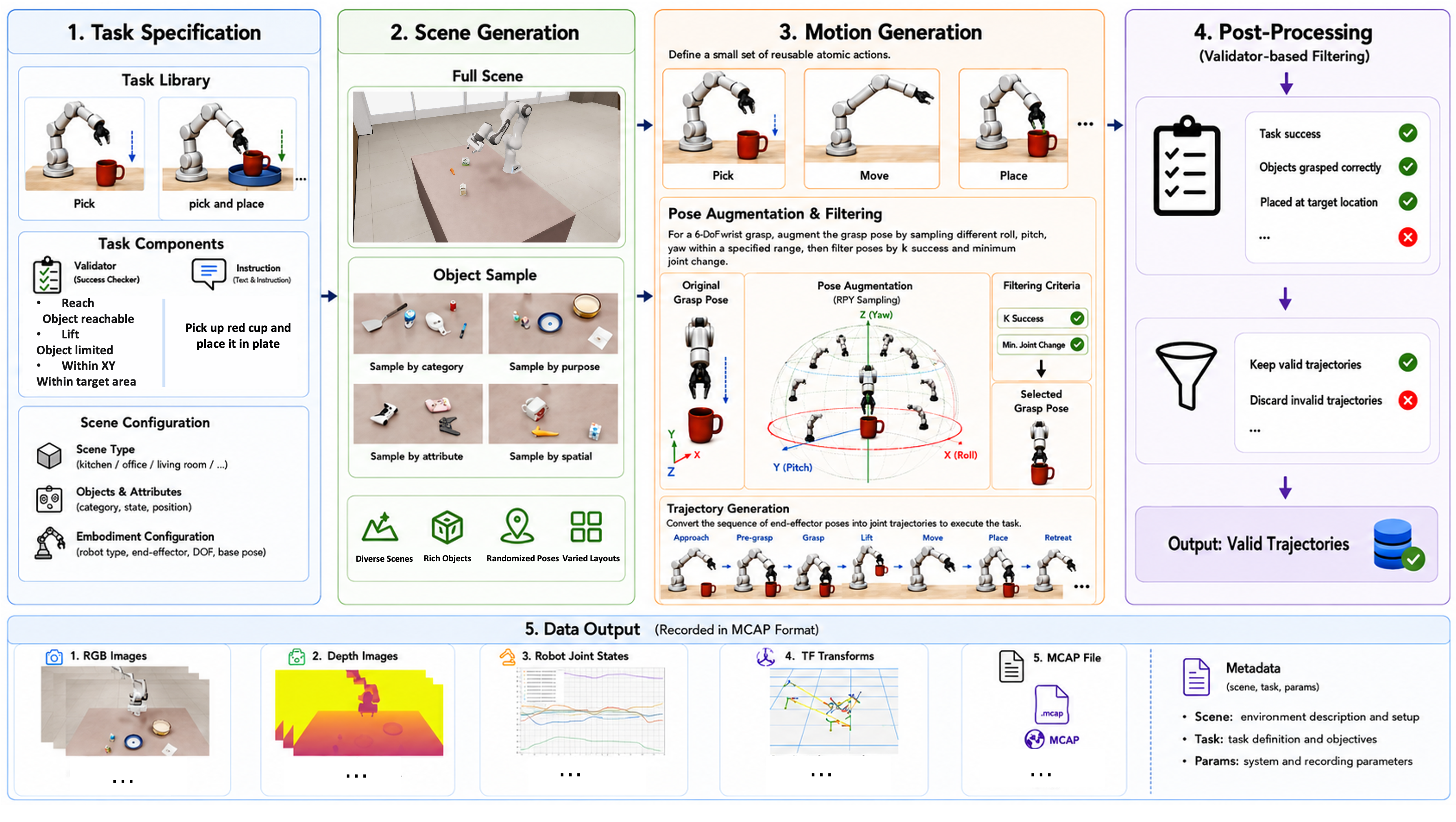}
\caption{Overview of the benchmark data generation pipeline, including Task Specification, Scene Generation, Motion Generation, Validator-Based Post-Processing, and Recorded Data Output.}
\label{fig:supp_data_pipeline}
\end{figure}

This appendix expands the infrastructure summarized in the main paper. The main benchmark definition only needs the high-level pipeline; Figure~\ref{fig:supp_data_pipeline} describes how \ours{} keeps task construction, data generation, and evaluation tied to the same episode specification.

\paragraph{Task Specification.}
A task specification defines what the robot should do and how success is measured. Each episode starts from a task specification, which serves as the shared source of truth for scene construction, motion generation, and evaluation. As shown in the first stage of Figure~\ref{fig:supp_data_pipeline}, the specification contains three parts: a scene configuration, a task library entry, and task components.

The scene configuration defines the execution context, including the background environment, object set and attributes, and robot embodiment. In the current benchmark, \ours{} supports multiple background environments and six embodiment configurations, allowing the same task logic to be instantiated under different visual and kinematic conditions.

The task library defines the reusable manipulation logic. We use two primitive templates, \textsc{Pick} and \textsc{Pick-and-Place}, and instantiate them into four benchmark task families: Pick Category, Pick Attribute, Pick Spatial, and Compositional Pick-Place. The first two emphasize semantic and attribute grounding, while the latter two emphasize spatial and compositional reasoning.

The task components specify how each episode is interpreted and evaluated. They include concrete objects, initial states, target conditions, natural-language instructions, and validator-based success criteria such as reachability, lifting, stability, collision freedom, and target-region placement. This organization separates reusable task logic from episode-specific objects and layouts, while keeping generation and evaluation aligned with the same specification.

\paragraph{Scene Generation.}
Scene generation instantiates a task specification into an executable simulation scene. Given the specified task family, object roles, success conditions, and robot embodiment, the pipeline constructs a concrete manipulation scene with selected object instances, initial object poses, spatial layouts, and physical properties.

Objects are sampled along several dimensions, including category, task purpose, object attributes, and spatial placement. This sampling strategy increases both visual and semantic diversity while ensuring that each generated episode remains consistent with the task specification and its validators. We provide the full asset curation and sampling procedure in Section~\ref{sec:supp_asset_pipeline}.

The scene generator further randomizes layouts, object poses, physical properties, and object combinations. These variations expose policies to a broader set of valid task configurations, improving robustness and supporting generalization across different scenes and object arrangements.

\paragraph{Motion Generation.}
The motion generation module converts high-level task specifications into physically executable robot trajectories. It first decomposes each manipulation task into a small set of reusable atomic actions, such as \textsc{Pick}, \textsc{Move}, and \textsc{Place}. These actions provide a structured representation of manipulation behavior. They also make trajectory synthesis reusable across tasks and embodiments.

For grasp-related actions, a single grasp pose may not satisfy the robot's kinematic constraints. Starting from an initial grasp pose, the module samples candidate poses by perturbing roll, pitch, and yaw within predefined ranges. It then filters these candidates by inverse-kinematics feasibility and joint-space cost. The first criterion ensures that the pose is reachable. The second favors solutions that require smaller changes from the current robot configuration. The selected pose is used as the final executable grasp target.

After selecting valid end-effector targets, the module converts the pose sequence into joint-space trajectories. Motion planning is performed under kinematic and collision constraints. The resulting trajectories are smooth, executable, and suitable for simulation rollout.

\paragraph{Post-Processing.}
Not all generated trajectories are suitable for the final dataset. We therefore apply an automatic validation stage after trajectory generation. Using the validators defined in the task specification, each rollout is checked against task-specific success criteria, such as correct grasping, object lifting, accurate placement, and overall task completion. Only trajectories that satisfy all required conditions are retained. Invalid rollouts are discarded, which improves the quality and reliability of the generated dataset. Across the four task instances, the valid synthesis rate is approximately 70\% for both Franka and PiperX embodiments.

\paragraph{Data Output.}
For each validated trajectory, the pipeline records multimodal observations and robot states in MCAP format. The exported data include RGB images, depth images aligned with the RGB frames, robot joint states, coordinate-frame transformations, and camera extrinsics. The joint states contain time-synchronized joint positions, velocities, and efforts. The TF records describe transformations among robot, camera, object, and scene frames, which supports geometry-aware analysis and reproducible playback.

MCAP is a container format for synchronized robotics data streams. It stores sensor observations and robot states in a single episode file for efficient loading, replay, and analysis. Each episode also includes metadata such as the scene configuration, task definition, recording parameters, and system configuration. This metadata supports reproducible evaluation and downstream dataset management.

\subsection{Asset Curation and Management Pipeline}
\label{sec:supp_asset_pipeline}

\begin{figure}[h]
\centering
\includegraphics[width=\linewidth]{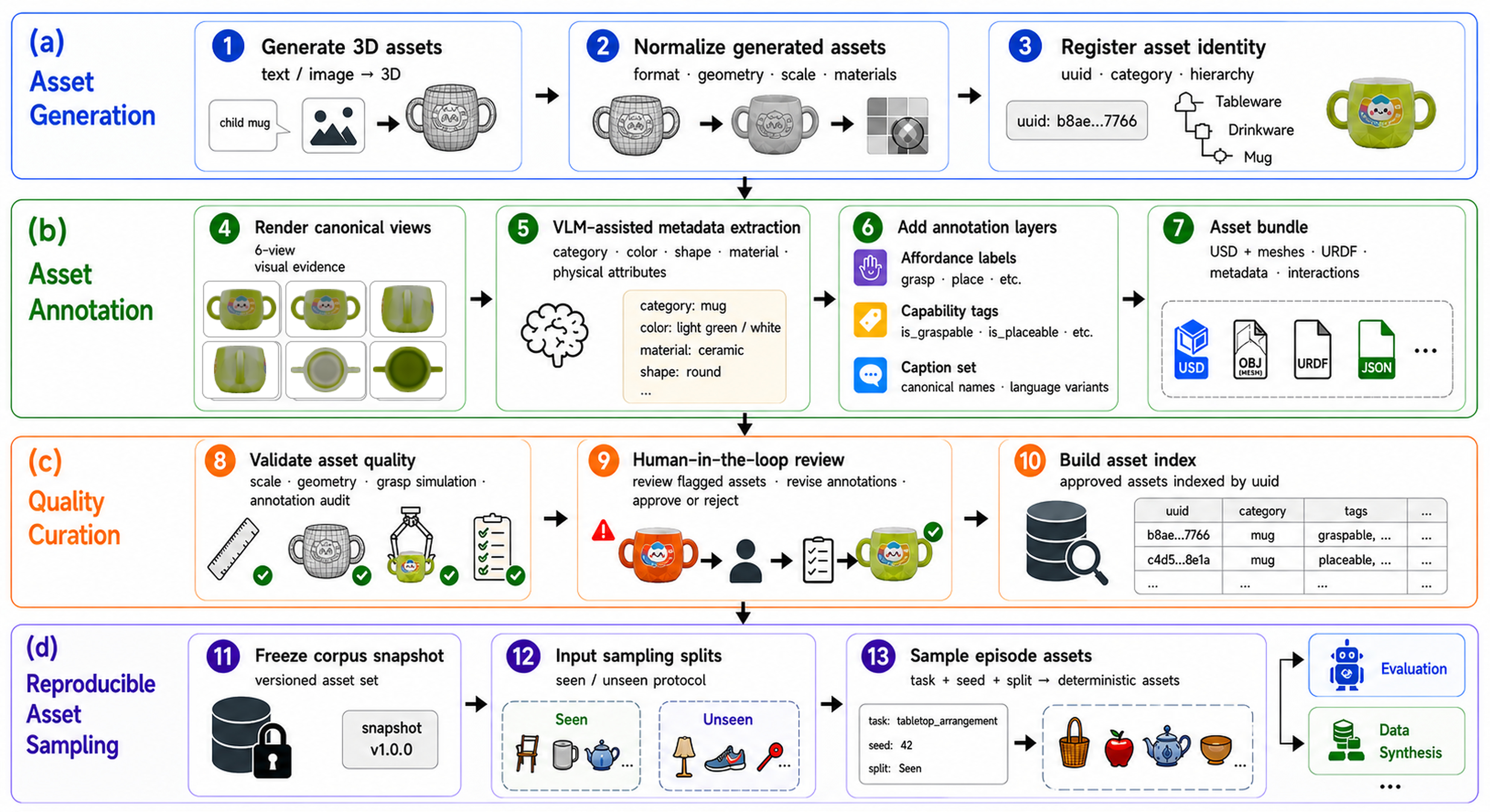}
\caption{Automated asset curation and management pipeline: (a)~asset generation, (b)~VLM-based annotation, (c)~quality curation with human-in-the-loop review, and (d)~reproducible sampling.}
\label{fig:asset_pipeline}
\end{figure}

\begin{figure}[h!]
\centering
\includegraphics[width=\linewidth]{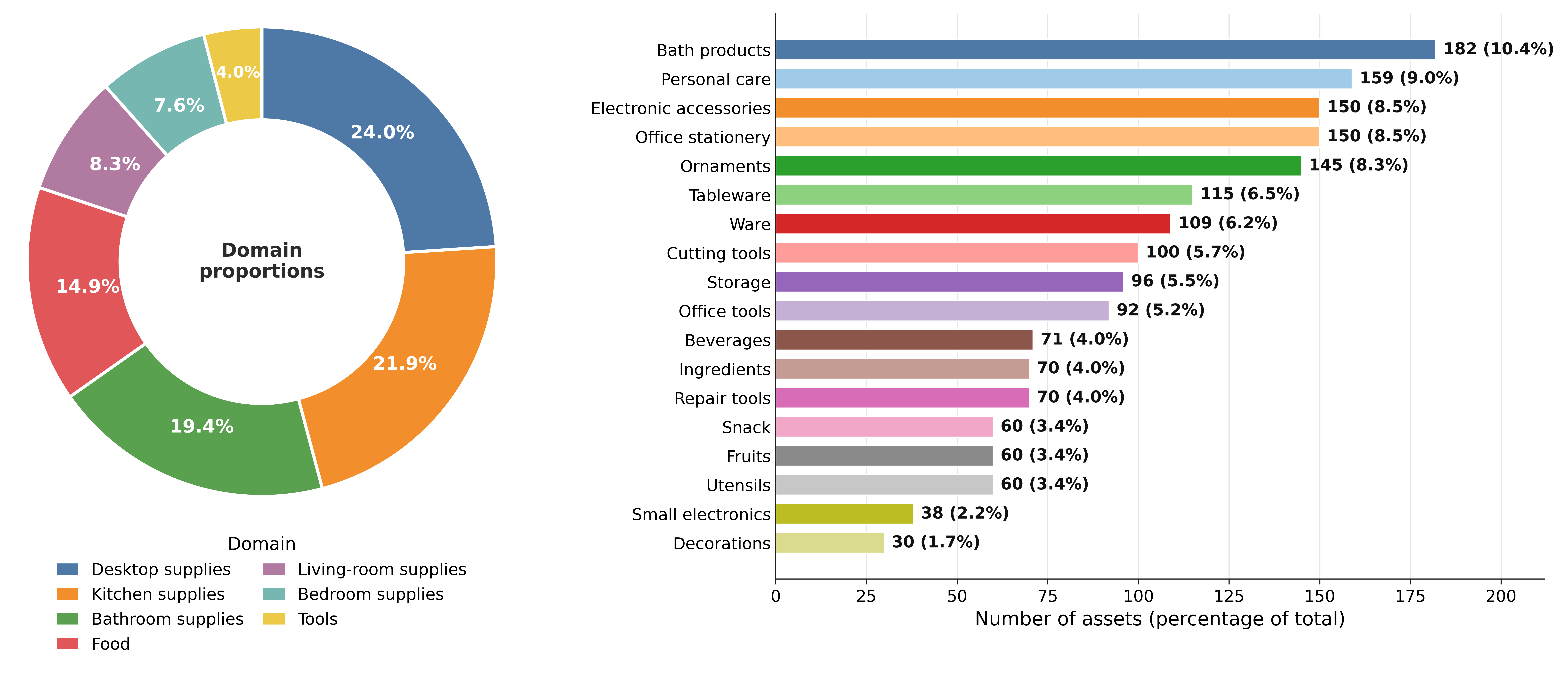}
\caption{Distribution of the asset corpus used in the benchmark and experiments. \textbf{Left}: domain proportions. \textbf{Right}: asset count per super-category (18 in total).}
\label{fig:asset_distribution}
\end{figure}

Figure~\ref{fig:asset_pipeline} illustrates the four-stage pipeline. The entire pipeline is automated end-to-end---from raw 3D generation through annotation and quality checks to reproducible sampling---with human review applied only to assets flagged during quality curation. The resulting corpus contains 1,757 objects spanning 7 domains, 18 super-categories, and 103 fine-grained categories (Figure~\ref{fig:asset_distribution}).

\paragraph{Asset generation.}
Raw 3D assets are produced by EmbodiedGen~\citep{wang2025embodiedgengenerative3dworld}, which synthesizes meshes from text or image prompts and exports USD scene descriptions, OBJ/GLB meshes, URDF articulation stubs, and PBR textures at estimated metric scale. Each asset is normalized to a canonical coordinate frame and registered under a four-level taxonomy (domain / super-category / category / instance) that defines both the semantic granularity of benchmark tasks and the axes for generalization splits. A persistent UUID is assigned at registration and remains fixed regardless of subsequent library reorganization.

\paragraph{Asset annotation.}
Six canonical views (front, back, left, right, top, bottom) are rendered and submitted to a VLM (GPT-4o), which returns structured metadata in a single query: super-category, category, free-form name and description, dominant color, shape class, material class, and plausible height and mass ranges. Categorical attributes are drawn from controlled vocabularies to keep metadata machine-filterable. On top of VLM-predicted attributes, we annotate three additional layers: (i)~\emph{manipulation affordances}---feasible grasp poses enumerated over canonical approach directions and filtered by a gripper-aperture feasibility criterion, plus gravity-aligned stable placement orientations; (ii)~\emph{capability tags} such as \texttt{is\_graspable} and \texttt{is\_container}, used for coarse task-level filtering; and (iii)~\emph{referring expressions}---short, action-agnostic noun phrases (e.g., ``the red ceramic mug'') partitioned into seen and unseen sets for language-generalization evaluation. These expressions serve both instruction generation and distractor sampling: they provide natural referring terms for targets and semantic cues for selecting plausible but instruction-inconsistent alternatives. All metadata is bundled into the asset's URDF record, making each object a self-describing unit.

\paragraph{Quality curation.}
We target two error modes that task success rate alone cannot detect. \emph{Scale errors}: because the simulator loads meshes at face value, a mis-scaled object is indistinguishable from a correctly modeled object of a different size---a policy can still grasp and place it, leaving absolute scale unconstrained. We check each object's metric size against both the VLM-predicted range and category-level commonsense priors, and repair flagged assets by uniformly rescaling geometry, bounding boxes, and affordance coordinates. \emph{Grasp-feasibility errors}: for each asset, we run 100 trial grasps in an Isaac Sim environment, sweeping placement positions and randomized orientations on a tabletop workspace and executing grasps from annotated affordance poses. Assets with low grasp success rates are flagged for review or excluded from the corpus. Semantic metadata is further cross-validated by two independent models (GPT-5.4 and Claude Sonnet 4.6); disagreements and consistency conflicts are routed to human annotators who confirm, reject, or correct each flagged item. Only assets that pass scale, grasp, and consistency checks enter the final corpus.

\paragraph{Reproducible asset sampling.}
The validated corpus is managed through two orthogonal controls. An immutable \emph{snapshot} freezes the exact set of objects available to an experiment, while a \emph{split} partitions that set into seen and unseen subsets for training and evaluation. Tasks declare only role-level semantic constraints (e.g., category, attribute, capability); a seeded sampler deterministically resolves these constraints under the active snapshot and split, so the same seed reproduces the same episode assignment even if the library is later expanded or reorganized.

\paragraph{Asset corpus.}
Figure~\ref{fig:asset_distribution} summarizes the corpus composition: the left panel shows the proportion of asset instances across the 7 domains, while the right panel breaks down asset instances per super-category. Table~\ref{tab:asset_corpus} provides the full fine-grained category listing. The corpus covers 7 domains, 18 super-categories, and 103 fine-grained categories, with roughly 17 generated instances per category on average. In addition to graspable objects, the corpus includes 20 receptacle or destination categories for constructing pick-and-place tasks.

\setlength{\LTleft}{0pt}
\setlength{\LTright}{0pt}
\scriptsize
\begin{longtable}{@{}p{0.17\linewidth}@{\hspace{0.5em}}p{0.22\linewidth}@{\hspace{0.5em}}>{\raggedright\arraybackslash}p{0.49\linewidth}@{\hspace{0.5em}}r@{}}
\caption{Detailed asset category inventory. The count column reports the number of fine-grained categories within each super-category; object instances are summarized separately in Figure~\ref{fig:asset_distribution}.}
\label{tab:asset_corpus}\\
\toprule
\textbf{Domain} & \textbf{Super-Category} & \textbf{Fine-Grained Categories} & \textbf{\# Categories} \\
\midrule
\endfirsthead
\toprule
\textbf{Domain} & \textbf{Super-Category} & \textbf{Fine-Grained Categories} & \textbf{\# Categories} \\
\midrule
\endhead
\midrule
\multicolumn{4}{r}{Continued on next page} \\
\endfoot
\bottomrule
\endlastfoot
\texttt{desktop\_supplies} & \texttt{office\_stationery} & \texttt{stapler}, \texttt{ballpoint\_pen}, \texttt{pen\_holder}, \texttt{glue\_stick}, \texttt{eraser}, \texttt{correction\_tape}, \texttt{marker}, \texttt{sharpener}, \texttt{notebook} & 9 \\
\texttt{desktop\_supplies} & \texttt{office\_tools} & \texttt{scissors}, \texttt{scotch\_tape}, \texttt{magnifier}, \texttt{cutter\_knife}, \texttt{stamp}, \texttt{staples\_box}, \texttt{sticky\_notes} & 7 \\
\texttt{desktop\_supplies} & \texttt{electronic\_accessories} & \texttt{mouse}, \texttt{tv\_remote}, \texttt{ac\_remote}, \texttt{presenter\_clicker}, \texttt{gamepad}, \texttt{earbud\_case}, \texttt{charging\_head}, \texttt{charging\_cable}, \texttt{usb\_drive} & 9 \\
\texttt{desktop\_supplies} & \texttt{decorations} & \texttt{mini\_sculpture}, \texttt{phone\_stand}, \texttt{vase} & 3 \\
\texttt{kitchen\_supplies} & \texttt{cutting\_tools} & \texttt{fruit\_knife}, \texttt{peeler}, \texttt{garlic\_press}, \texttt{wire\_cutters} & 4 \\
\texttt{kitchen\_supplies} & \texttt{tableware} & \texttt{bowl}, \texttt{chopsticks}, \texttt{coaster}, \texttt{fork}, \texttt{plate}, \texttt{spoon}, \texttt{tray} & 7 \\
\texttt{kitchen\_supplies} & \texttt{ware} & \texttt{wine\_glass}, \texttt{mug}, \texttt{coffee\_cup}, \texttt{thermos}, \texttt{baby\_bottle} & 5 \\
\texttt{kitchen\_supplies} & \texttt{utensils} & \texttt{spice\_jar}, \texttt{bottle\_opener}, \texttt{spatula} & 3 \\
\texttt{bedroom\_supplies} & \texttt{storage} & \texttt{storage\_box}, \texttt{jewelry\_box}, \texttt{glasses\_case}, \texttt{pill\_organizer}, \texttt{remote\_holder} & 5 \\
\texttt{bedroom\_supplies} & \texttt{small\_electronics} & \texttt{flashlight}, \texttt{alarm\_clock} & 2 \\
\texttt{bathroom\_supplies} & \texttt{personal\_care} & \texttt{toothbrush}, \texttt{toothpaste}, \texttt{mouthwash\_cup}, \texttt{comb}, \texttt{razor}, \texttt{nail\_clipper}, \texttt{dental\_floss}, \texttt{hair\_clip} & 8 \\
\texttt{bathroom\_supplies} & \texttt{bath\_products} & \texttt{shampoo\_bottle}, \texttt{soap\_dish}, \texttt{lotion}, \texttt{hand\_sanitizer}, \texttt{sponge}, \texttt{facial\_cleanser} & 6 \\
\texttt{livingroom\_supplies} & \texttt{ornaments} & \texttt{houseplant}, \texttt{watering\_can}, \texttt{diffuser\_bottle}, \texttt{photo\_frame}, \texttt{candle\_holder} & 5 \\
\texttt{tools} & \texttt{repair\_tools} & \texttt{screwdriver}, \texttt{wrench}, \texttt{hex\_key}, \texttt{measuring\_tape}, \texttt{pliers}, \texttt{mini\_hammer}, \texttt{glue\_gun} & 7 \\
\texttt{food} & \texttt{fruits} & \texttt{apple}, \texttt{lemon}, \texttt{banana}, \texttt{pear}, \texttt{strawberry}, \texttt{peach} & 6 \\
\texttt{food} & \texttt{ingredients} & \texttt{potato}, \texttt{tomato}, \texttt{egg}, \texttt{bread}, \texttt{garlic}, \texttt{carrot}, \texttt{chili} & 7 \\
\texttt{food} & \texttt{beverages} & \texttt{juice}, \texttt{milk\_carton}, \texttt{canned\_coke}, \texttt{beverage\_bottle} & 4 \\
\texttt{food} & \texttt{snack} & \texttt{candy}, \texttt{chewing\_gum}, \texttt{biscuit}, \texttt{potato\_chips}, \texttt{chocolate}, \texttt{cheese\_sticks} & 6 \\
\midrule
\multicolumn{3}{r}{\textbf{Total}} & \textbf{103} \\
\end{longtable}
\normalsize

\subsection{Instruction Generation and Language Diversity}
\label{sec:supp_instruction_diversity}

\ours{} constructs natural-language instructions by composing task-level paraphrases with object-level referring expressions. This design varies the surface form of an instruction while preserving its underlying task semantics and instruction-consistent referents.

\paragraph{Template-level paraphrases.}
Each task family registers a canonical instruction template together with a pool of semantically equivalent variants. The templates contain object-description slots that are instantiated when an episode is generated. For example, the Compositional Pick-Place task uses variants such as ``Pick up \{actor1.description\} and place in \{actor2.description\},'' ``Grab \{actor1.description\} and place it in \{actor2.description\},'' ``Grasp \{actor1.description\} and set it inside \{actor2.description\},'' and ``Lift \{actor1.description\} and drop it into \{actor2.description\}.'' The variant pool covers diverse manipulation verbs, placement verbs, and spatial prepositions while keeping the intended action unchanged.

\paragraph{Object-level referring expressions.}
For each object asset, a large language model generates multiple caption candidates that describe discriminative visual and semantic properties such as category, color, shape, material, and geometry. During episode generation, a caption candidate is sampled for each object slot in the selected template.

By combining template-level paraphrases with object-level caption candidates, the same underlying task instance can be expressed through many linguistically distinct instructions. This compositional generation process increases instruction diversity without changing the task goal or the identity of the referred objects.

\subsection{Domain Randomization}
\label{sec:supp_domain_randomization}

\begin{figure}[h!]
\centering
\includegraphics[width=\linewidth]{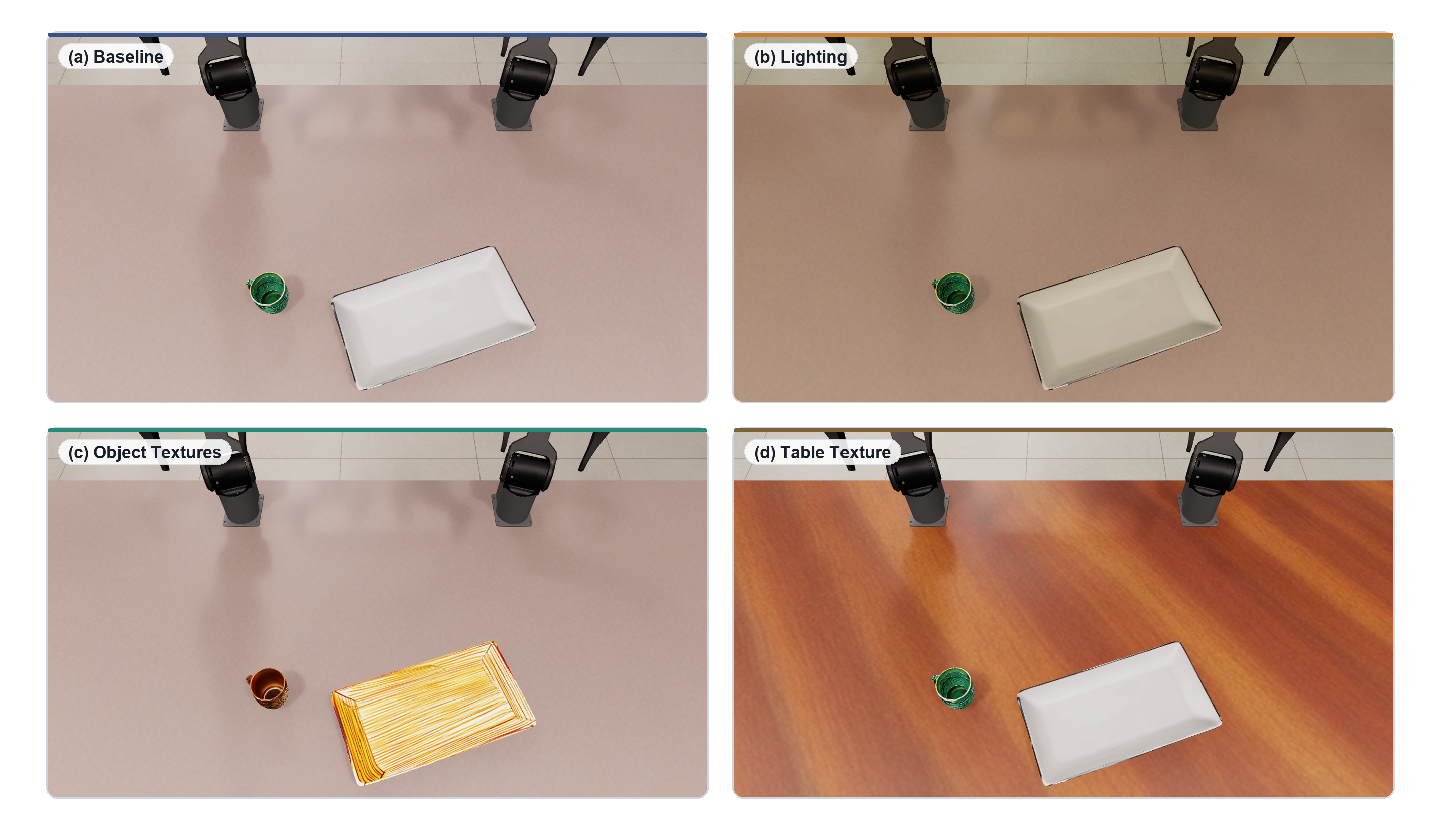}
\caption{Domain randomization on a fixed manipulation scene. (a)~Baseline with default lighting and textures. (b)~Randomized illumination. (c)~Randomized object surface textures. (d)~Randomized table surface material. Scene layout and embodiment are held fixed so that each panel differs in a single factor.}
\label{fig:domain_randomization}
\end{figure}

To improve visual robustness and facilitate sim-to-real transfer, \ours{} exposes domain randomization (DR) as a declaratively configured component of the task specification. Each task declares the randomization factors to activate in its configuration, and the corresponding stochastic transforms are applied at every environment reset as seeded event terms. Sharing the random seed across episodes fixes object placement and embodiment configuration, so that any individual factor can be isolated for controlled, reproducible comparison (Figure~\ref{fig:domain_randomization}).

We randomize two complementary modalities. \emph{Photometric randomization} perturbs scene illumination: the correlated color temperature, tint, and intensity of each light source are resampled within configurable ranges. Because scenes are composed from referenced photorealistic indoor environments with embedded area, dome, and distant emitters, the randomizer recursively traverses the asset subtree and jointly perturbs all contained lights, producing globally coherent yet substantially varied lighting conditions. \emph{Appearance randomization} alters surface materials. Every manipulable object, receptacle, and table is authored with a USD VariantSet enumerating physically plausible material realizations---from categorical finishes per object (e.g., ceramic, brushed metal, glazed) to procedurally generated surface textures per table. At reset, the randomizer samples a variant per asset; the same mechanism supports deterministic pinning of a named variant for controlled appearance sweeps and ablations.

\subsection{Real-World Experimental Setup}
\label{sec:supp_real_world_setup}

\begin{figure}[h!]
\centering
\begin{minipage}[c]{0.48\linewidth}
\centering
\includegraphics[width=\linewidth]{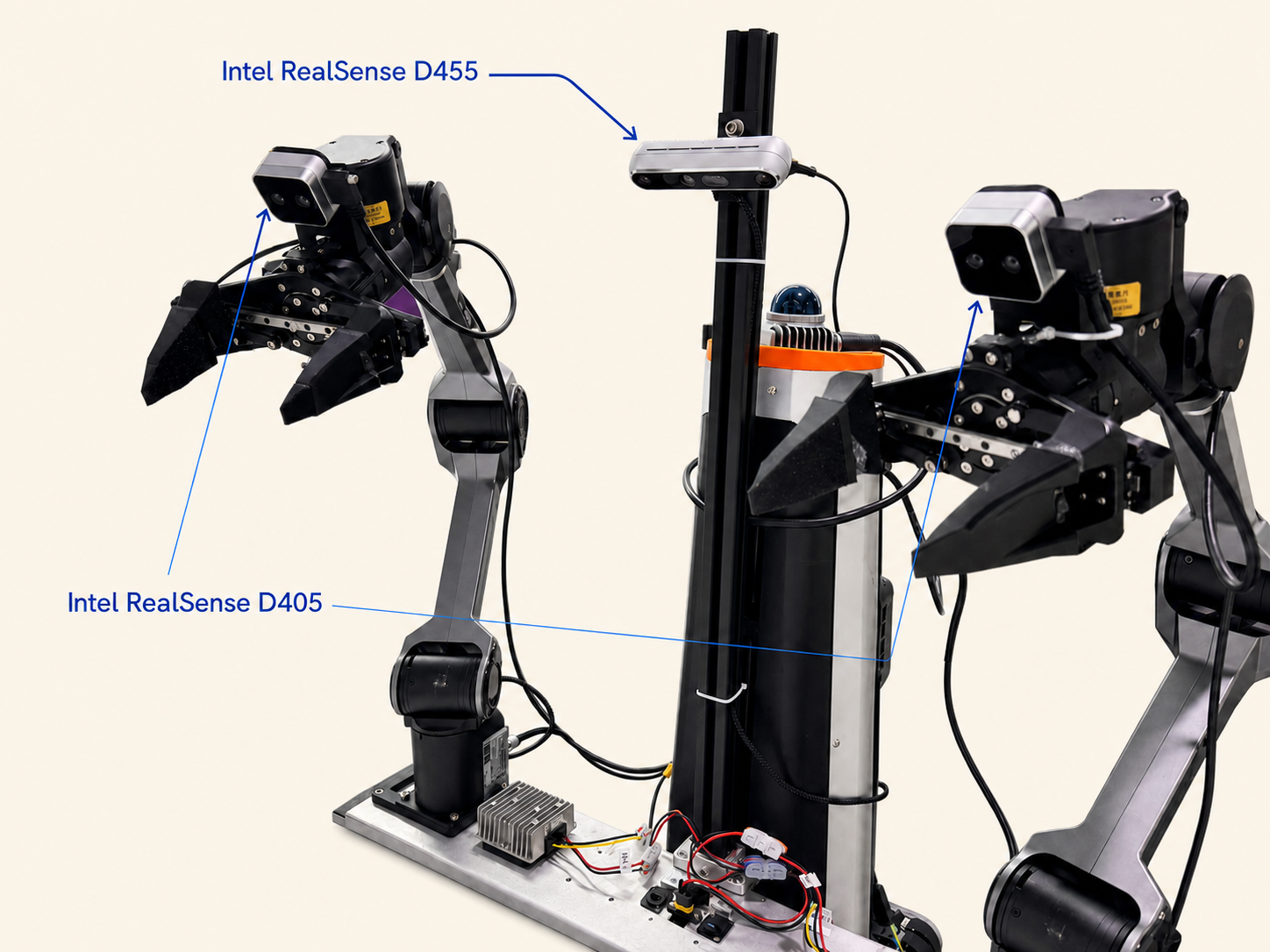}
\end{minipage}
\hfill
\begin{minipage}[c]{0.48\linewidth}
\centering
\includegraphics[width=\linewidth]{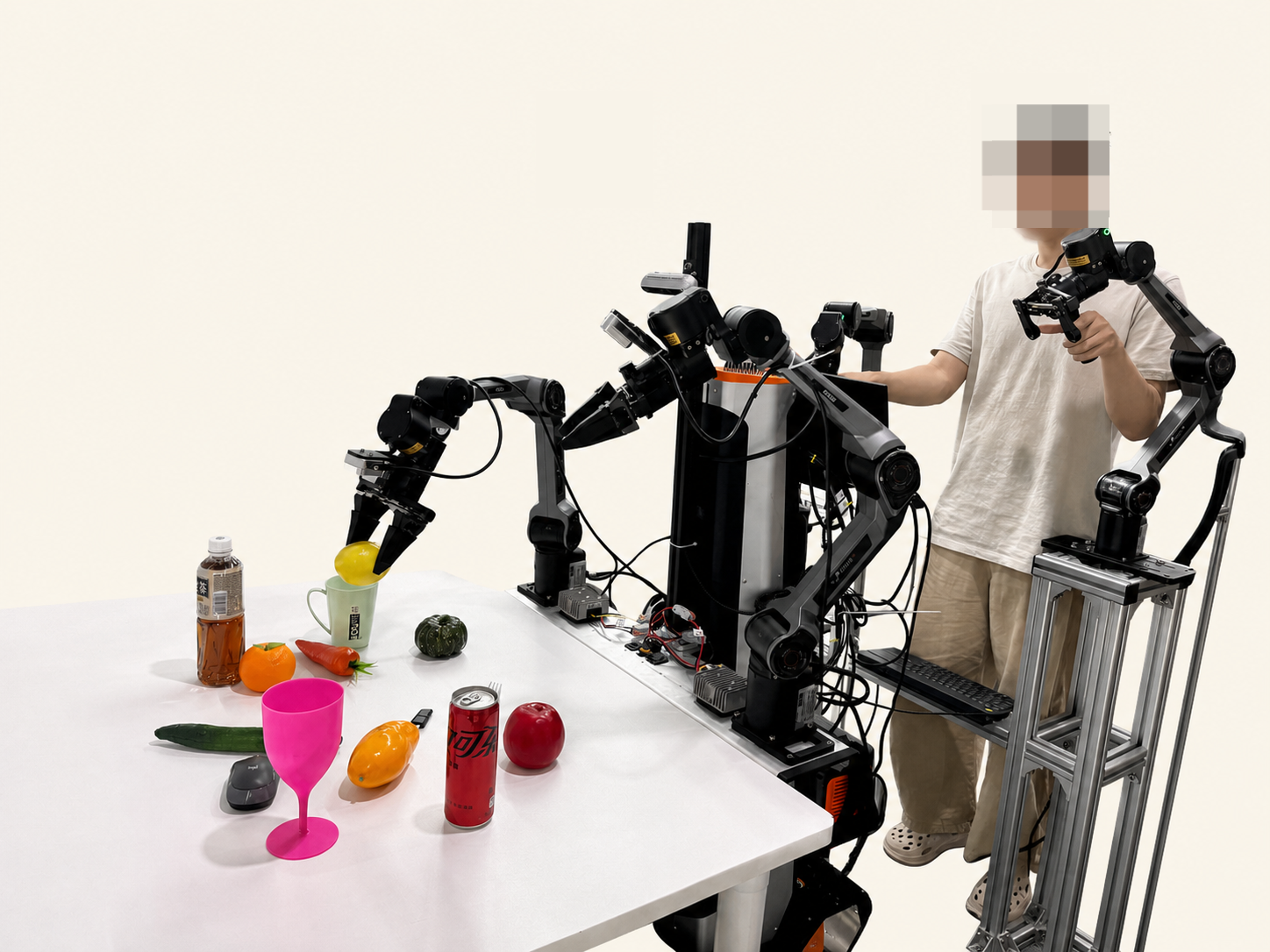}
\end{minipage}
\caption{Real-world experimental setup. \textbf{Left}: dual-arm AgileX PiperX robot with one static camera and two wrist-mounted cameras. \textbf{Right}: ALOHA teleoperation setup for data collection.}
\label{fig:real_exp_setting}
\end{figure}

We conduct real-world experiments on a dual-arm AgileX PiperX robot (Figure~\ref{fig:real_exp_setting}) equipped with three RGB-D cameras: one Intel RealSense D455 mounted directly above the workspace as a global view, and two Intel RealSense D405 attached to the left and right wrists respectively. All tasks involve fixed-base tabletop manipulation. Data collection uses homologous teleoperation following the ALOHA methodology~\citep{zhao2023aloha}, where an operator controls two 6-DoF slave arms to demonstrate each task.

\section{$\pi_{0.5}$ Target Binding Analysis}
\label{sec:supp_pi05_grounding}

\subsection{Failure Case Analysis}
\label{sec:supp_pi05_failure}

We analyze the 18 Reach-failed episodes of $\pi_{0.5}$ on the \texttt{pick\_category} seen split. The Reach metric is instruction-conditioned, so acting on an incorrect object is counted as a failure even when the low-level motion succeeds. Under this criterion, most failed cases still complete a lift on a distractor object.

\begin{figure}[htbp]
    \centering
    \includegraphics[width=\linewidth]{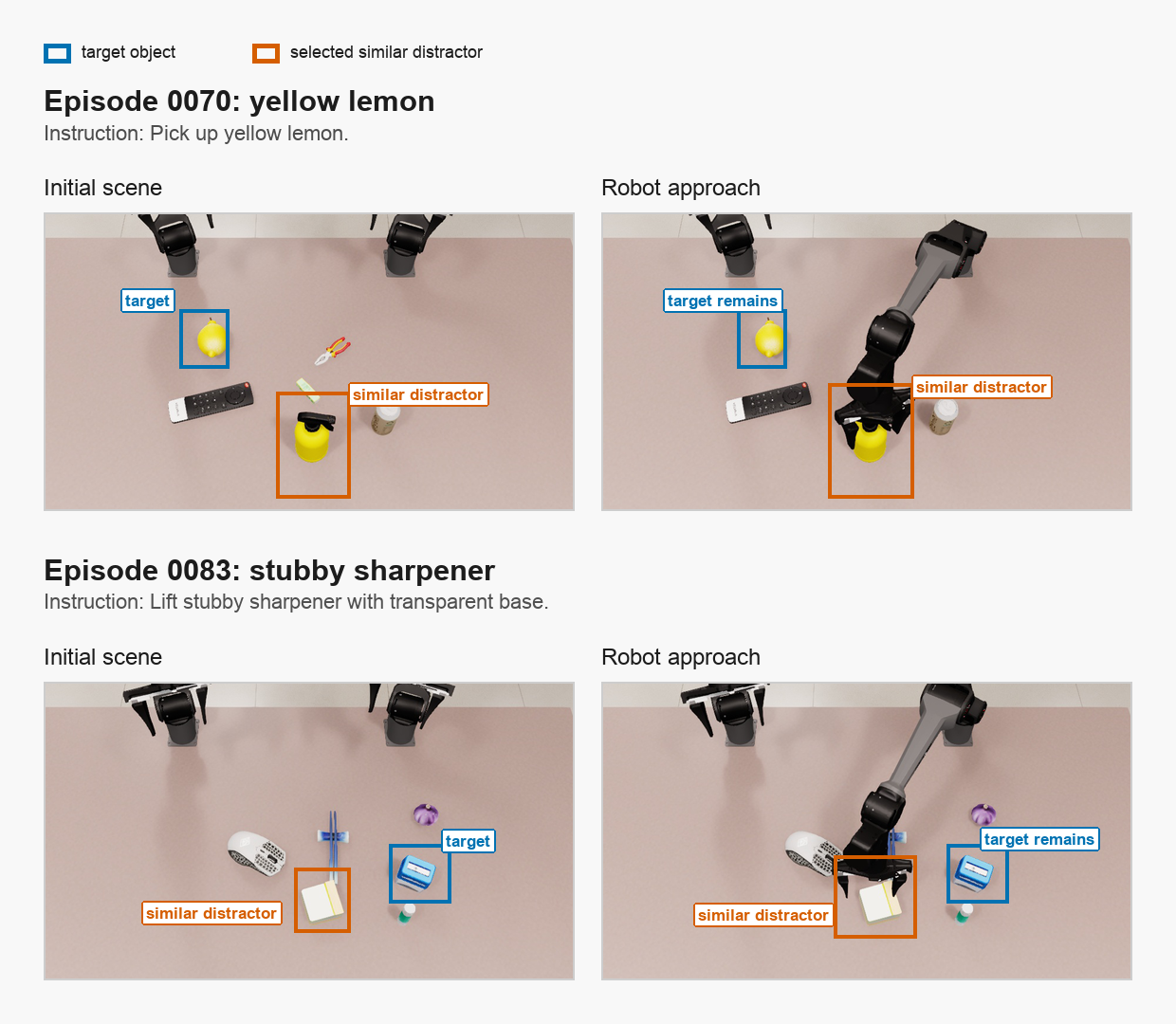}
    \caption{Two representative Reach-failed $\pi_{0.5}$ episodes on the \texttt{pick\_category} seen split. Blue boxes mark the instruction target; orange boxes mark the similar distractor selected by the policy. The right panels show the robot approaching the distractor.}
    \label{fig:pi05_reach_failure_examples}
\end{figure}

These failures are therefore not missing-action or low-level grasp failures. The policy executes a physically valid pick, but on an instruction-inconsistent object. Figure~\ref{fig:pi05_reach_failure_examples} shows two representative cases: the policy approaches a yellow lotion bottle for ``yellow lemon'' and a cube-shaped distractor for ``stubby sharpener.'' Across the 18 Reach failures, 15 lifted distractors share salient cues with the target, such as color, shape, geometry, or affordance.

This pattern suggests partial grounding without reliable referent binding: $\pi_{0.5}$ sometimes responds to one salient attribute in the instruction, such as color or shape, but does not reliably bind the full referring expression to the intended object category. This failure mode explains why text-indispensable distractors are necessary. In simpler scenes, partial grounding may still succeed when the visually matching object is also the target. In \ours{}, multiple feasible candidates can share salient cues, so the policy must ground the full instruction rather than select a plausible partial match.

\section{Training Recipes}
\label{sec:supp_training_recipes}

This section documents the training recipes of the four evaluated policies: $\pi_0$, $\pi_{0.5}$, GR00T N1.7, and Motus. Because the policies differ in architecture, pretraining, and embodiment adaptation, we first describe the shared protocol and then present model-specific settings. Unless stated otherwise, all models are trained for 5{,}000 steps and evaluated at the final checkpoint; no checkpoint selection is performed.

\subsection{Shared Training Protocol}
\label{subsec:supp_shared_training}

All models are trained under the benchmark protocol described in the main paper. For the dual-arm PiperX, training covers \texttt{pick\_category}, \texttt{pick\_attribute}, and \texttt{place\_a2b}. For the single-arm Franka, training covers \texttt{pick\_spatial} and \texttt{place\_a2b}. Each model uses its own observation preprocessing and action interface, as the underlying architectures do not share a common tokenization or control representation. All policies are fine-tuned on benchmark demonstrations using their native training pipelines; model-specific adaptations are detailed below.

\subsection{Model-Specific Recipes}
\label{subsec:supp_model_recipes}

Table~\ref{tab:supp_training_recipe_summary} provides an at-a-glance summary; detailed descriptions follow.

\begin{table}[H]
\centering
\caption{Summary of model-specific training recipes. Unlogged details are qualified in the text below.}
\label{tab:supp_training_recipe_summary}
\begin{tabular}{l l l l}
\toprule
\textbf{Model} & \textbf{Init checkpoint} & \textbf{Trainable scope} & \textbf{Batch / Steps} \\
\midrule
$\pi_0$ & OpenPI \texttt{pi0\_base} & LoRA only & 32/GPU, 5k \\
$\pi_{0.5}$ & OpenPI \texttt{pi05\_base} & LoRA only & 32/GPU, 5k \\
GR00T N1.7 & \texttt{GR00T-N1.7-3B} & Proj./diff./VL-norm & 24/GPU, 5k \\
Motus & \texttt{motus-robotics/Motus} & Qwen3-VL frozen & 8/GPU, 5k \\
\bottomrule
\end{tabular}
\end{table}

\paragraph{$\pi_0$ and $\pi_{0.5}$.}
Both policies share the same LoRA-based fine-tuning recipe and differ only in their initialization checkpoint: $\pi_0$ uses the OpenPI \texttt{pi0\_base} checkpoint, while $\pi_{0.5}$ uses \texttt{pi05\_base}. The configuration employs \texttt{gemma\_2b\_lora} for the PaliGemma vision-language backbone and \texttt{gemma\_300m\_lora} for the action expert, with all non-LoRA parameters frozen. The model consumes RGB observations resized to $224 \times 224$, the current robot state, and a tokenized language instruction. The benchmark adapter exposes robot-specific action interfaces: PiperX uses a 14-dimensional dual-arm joint-action chunk, and Franka uses an 8-dimensional single-arm interface. Both predict 50-step action chunks. Training uses AdamW with cosine learning-rate decay, 1{,}000 warmup steps, a peak learning rate of $2.5 \times 10^{-5}$, gradient clipping at $1.0$, no EMA, a per-GPU batch size of 32, and 5{,}000 steps on 8 NVIDIA RTX 5090 GPUs. Checkpoints are saved every 1{,}000 steps, and the reported results use the final 5{,}000-step checkpoint. Training takes approximately 2.5 hours. In the PiperX sim-to-real transfer setting, a 9:1 batch-sampling ratio is applied between simulation and real-world data.

\paragraph{GR00T N1.7.}
GR00T N1.7 is fine-tuned via robot-specific AnyMove runs initialized from a local mirror of the official \texttt{GR00T-N1.7-3B} checkpoint, with the VLM backbone resolved to a local \texttt{Cosmos-Reason2-2B} path. The benchmark adaptation registers custom \texttt{NEW\_EMBODIMENT} modality configurations for both robot settings. Both use three RGB cameras, proprioceptive state, and language instructions. The Franka setting uses \texttt{ext1\_camera}, \texttt{ext2\_camera}, and \texttt{wrist\_camera}; PiperX uses \texttt{static\_camera}, \texttt{left\_hand\_camera}, and \texttt{right\_hand\_camera}. The action interface is joint-space: Franka uses relative joint actions with absolute gripper commands, and PiperX uses relative left/right arm-joint actions with absolute gripper commands. The embodiment exposes 30 valid action steps within a 40-step model horizon, with the unused tail masked. Training freezes the LLM and visual backbone while updating the projector, diffusion, and VL-normalization modules. The optimizer is AdamW with cosine decay, a warmup ratio of 0.05, BF16, gradient clipping at $1.0$, DeepSpeed ZeRO-3, and gradient checkpointing. The completed artifacts correspond to 5{,}000-step runs at global batch size 192 on 8 GPUs. Training runs without an intermediate validation loop, and the reported results use the final 5{,}000-step checkpoint.

\paragraph{Motus.}
Motus is fine-tuned through experiment-specific configurations under \texttt{cluster/motus/}, initialized from the \texttt{motus-robotics/Motus} checkpoint. When this checkpoint is present, the training entrypoint skips independent WAN and VLM pretrained loading and partially loads the Motus weights directly. The observation pipeline produces a three-camera T-shaped stitched RGB image resized to $384 \times 320$, together with the robot state, WAN T5 language embeddings (text length 512), and Qwen3-VL image-text inputs. The action head predicts continuous normalized action chunks of length 48, corresponding to 8 video frames with a video-to-action ratio of 6. The embodiment interface is robot-specific: PiperX uses 14-dimensional state/action vectors covering \texttt{pick\_category}, \texttt{pick\_attribute}, and \texttt{place\_a2b}; Franka uses 8-dimensional vectors covering \texttt{pick\_spatial} and \texttt{place\_a2b}. Motus does not use LoRA; Qwen3-VL remains frozen while the remaining modules are optimized with AdamW ($\beta_1{=}0.9$, $\beta_2{=}0.95$), a learning rate of $10^{-5}$, weight decay of $0.01$, gradient clipping at $0.5$, BF16, 200 warmup steps, and a linear scheduler over 5{,}000 steps. The per-process batch size is 8, yielding an effective batch size of 64 under the 8-GPU setup. The reported results use the 5{,}000-step checkpoint.

\end{document}